%% file: root.tex
\documentclass[letterpaper, 10 pt, conference]{ieeeconf}  

\IEEEoverridecommandlockouts                              

\usepackage{times}
\usepackage{nicefrac}
\usepackage{myAmsthm}

\newtheoremstyle{mystyle}
  {}
  {}
  {\itshape}
  {}
  {\bfseries}
  {.}
  { }
  {\thmname{#1}\thmnumber{ #2}\thmnote{ (#3)}}%

\usepackage{multicol}
\usepackage[bookmarks=true,colorlinks=true,urlcolor=red,pagebackref]{hyperref}

\input{includes/usepackages}
\usepackage{caption}
\usepackage{subcaption}
\usepackage{float}
\newcommand{\supplementary}{appendix}

\input{includes/preamble_packages}

\input{includes/preamble_symbols}

\input{includes/preamble_algo}

\input{includes/shortcuts} 

\title{\LARGE \bf
Data-Association-Free Landmark-based SLAM
}

\author{Yihao Zhang$^{1}$, Odin A. Severinsen$^{1}$, John J. Leonard$^{1}$, Luca Carlone$^{1}$,  Kasra Khosoussi$^{2}$
\thanks{*This work was partially funded by ARL DCIST CRA
W911NF-17-2-0181, Lincoln Laboratory's Resilient
Perception in Degraded Environments program, ONR Neuroautonomy MURI grant N00014-19-1-2571, and ARPA-E award DE-AR0001218 under the
DIFFERENTIATE Program.}
\thanks{$^{1}$Y.\,Zhang, O.\,Severinsen, J.\,Leonard, and L.\,Carlone are with the Massachusetts Institute of Technology, United States,
        {\tt\footnotesize \{yihaozh,odinase,jleonard,lcarlone\}@mit.edu}}%
\thanks{$^{2}$K.\,Khosoussi is with the Commonwealth Scientific and Industrial Research Organisation (CSIRO), Australia,
        {\tt\footnotesize Kasra.Khosoussi@data61.csiro.au}}%
}

\begin{document}
\def\Name{Data-Association-Free Landmark-Based SLAM}

\maketitle
\thispagestyle{empty}
\pagestyle{empty}

\setcounter{figure}{1}

\input{sections/abstract}

\IEEEpeerreviewmaketitle

\input{sections/intro}
\input{sections/relatedWorks}
\input{sections/problemFormulations}

\input{sections/solvers}
\input{sections/experiments}

\input{sections/conclusion}

\pagebreak
{\smaller 
\bibliographystyle{plain}
\bibliography{root.bib,includes/refs.bib}
}

\newpage
\clearpage

\input{sections/appendix}

\end{document}

%% file: includes/usepackages.tex
\newcommand\inputpgf[2]{{
\let\pgfimageWithoutPath\pgfimage
\renewcommand{\pgfimage}[2][]{\pgfimageWithoutPath[##1]{#1/##2}}
\input{#1/#2}
}}

\usepackage{tabularx}
\newcolumntype{Y}{>{\centering\arraybackslash}X} 

%% file: includes/preamble_packages.tex
\usepackage{comment}
\usepackage{siunitx}
\usepackage{relsize}
\usepackage{ifthen}
\usepackage[colorinlistoftodos]{todonotes}

\usepackage[vlined,ruled,linesnumbered]{algorithm2e}
\usepackage{graphics} 
\usepackage{rotating}
\usepackage{color}
\usepackage{enumerate}
\usepackage[T1]{fontenc}
\usepackage{psfrag}
\usepackage{epsfig} 
\usepackage{booktabs}
\usepackage{graphicx,url}
\usepackage{multirow}
\usepackage{array}
\usepackage{latexsym}
\usepackage{amsfonts}
\usepackage{amsmath}
\usepackage{amssymb}
\usepackage{xstring}
\usepackage[noend]{algorithmic}
\usepackage{multirow}
\usepackage{xcolor}
\usepackage{prettyref}
\usepackage{flexisym}
\usepackage{bigdelim}
\usepackage{breqn} 
\usepackage{listings}
\let\labelindent\relax
\usepackage{enumitem}
\usepackage{xspace}
\usepackage{bm}
\graphicspath{{./figures/}}
\usepackage{tikz}
\usetikzlibrary{matrix,calc}

\usepackage{mdwlist}

\let\itemize\undefined
\makecompactlist{itemize}{stditemize}


%% file: includes/preamble_symbols.tex
\newrefformat{prob}{Problem\,\ref{#1}}
\newrefformat{def}{Definition\,\ref{#1}}
\newrefformat{sec}{Section\,\ref{#1}}
\newrefformat{sub}{Section\,\ref{#1}}
\newrefformat{prop}{Proposition\,\ref{#1}}
\newrefformat{app}{Appendix\,\ref{#1}}
\newrefformat{alg}{Algorithm\,\ref{#1}}
\newrefformat{cor}{Corollary\,\ref{#1}}
\newrefformat{thm}{Theorem\,\ref{#1}}
\newrefformat{lem}{Lemma\,\ref{#1}}
\newrefformat{fig}{Fig.\,\ref{#1}}
\newrefformat{tab}{Table\,\ref{#1}}

\newtheorem{theorem}{Theorem}
\newtheorem{problem}{Problem}

\newtheorem{proposition}[theorem]{Proposition}
\newtheorem{remark}[theorem]{Remark}

\newcommand{\bdmath}{\begin{dmath}}
\newcommand{\edmath}{\end{dmath}}
\newcommand{\beq}{\begin{equation}}
\newcommand{\eeq}{\end{equation}}
\newcommand{\bdm}{\begin{displaymath}}
\newcommand{\edm}{\end{displaymath}}
\newcommand{\bea}{\begin{eqnarray}}
\newcommand{\eea}{\end{eqnarray}}
\newcommand{\beal}{\beq \begin{array}{ll}}
\newcommand{\eeal}{\end{array} \eeq}
\newcommand{\beas}{\begin{eqnarray*}}
\newcommand{\eeas}{\end{eqnarray*}}
\newcommand{\ba}{\begin{array}}
\newcommand{\ea}{\end{array}}
\newcommand{\bit}{\begin{itemize}}
\newcommand{\eit}{\end{itemize}}
\newcommand{\ben}{\begin{enumerate}}
\newcommand{\een}{\end{enumerate}}

\newcommand{\calM}{{\cal M}}

\newcommand{\eg}{\emph{e.g.,}\xspace}
\newcommand{\ie}{\emph{i.e.,}\xspace}
\newcommand{\myParagraph}[1]{{\bf #1.}\xspace}

\newcommand{\M}[1]{{\bm #1}} 
\renewcommand{\boldsymbol}[1]{{\bm #1}}

\newcommand{\hide}[1]{}

\newcommand{\hiddenText}{{\color{gray} hidden text.}}
\newcommand{\hideWithText}[1]{\hiddenText}

\newcommand{\Natural}[1]{ { {\mathbb N}^{#1} } }

\newcommand{\prob}[1]{{\mathbb P}\left(#1\right)}
\newcommand{\tran}{^{\mathsf{T}}}

\newcommand{\Real}[1]{ { {\mathbb R}^{#1} } }

\newcommand{\at}[1]{^{(#1)}}

\newcommand{\SE}[1]{\ensuremath{\mathrm{SE}(#1)}\xspace}

\newcommand{\MR}{\M{R}}

\newcommand{\vt}{\boldsymbol{t}}
\newcommand{\vxx}{\boldsymbol{x}} 
\newcommand{\vy}{\boldsymbol{y}}

\newcommand{\vepsilon}{\boldsymbol{\epsilon}}

\newcommand{\blue}[1]{{\color{blue}#1}}

\newcommand{\linkToPdf}[1]{\href{#1}{\blue{(pdf)}}}
\newcommand{\linkToPpt}[1]{\href{#1}{\blue{(ppt)}}}
\newcommand{\linkToCode}[1]{\href{#1}{\blue{(code)}}}
\newcommand{\linkToWeb}[1]{\href{#1}{\blue{(web)}}}
\newcommand{\linkToVideo}[1]{\href{#1}{\blue{(video)}}}
\newcommand{\linkToMedia}[1]{\href{#1}{\blue{(media)}}}
\newcommand{\award}[1]{\xspace} 

\newcommand{\vz}{\boldsymbol{z}}


%% file: includes/preamble_algo.tex
\lstdefinelanguage{mine} 
{morekeywords={while,True,if,break,=,return,function,for,until,in,input,output,assumptions,assumption,invariant,loop,variant,end,invariant,precondition,variables}, 
sensitive=false, 
morecomment=[l]{\#}, 
morecomment=[il]\%{.}, 
morecomment=[s]{/*}{*/}, 
morestring=[b]", 
} 

\usepackage{calc}
\newlength\listingnumberwidth

%% file: includes/shortcuts.tex
\newcommand{\probName}{DAF-SLAM\xspace} 

\newcommand{\ProbNameLong}{Data-Association-Free SLAM\xspace}
\newcommand{\vzbar}{\bar{\vz}} 
\newcommand{\vzhat}{\hat{\vz}}

\newcommand{\nameKmeans}{K-SLAM\xspace}

\newcommand{\nameAlt}{Oracle\xspace}
\newcommand{\nameML}{ML\xspace}
\newcommand{\nameOdom}{Odom\xspace}

%% file: sections/abstract.tex

\begin{abstract}
We study landmark-based SLAM with unknown data association: our robot navigates
in a completely unknown environment and has to simultaneously reason over its
own trajectory, the positions of an unknown number of landmarks in the environment, and
potential data associations between measurements and landmarks.  
This setup is interesting since: (i) it
arises when recovering from data association failures or from SLAM with
information-poor sensors, (ii) it sheds light on fundamental limits (and
hardness) of landmark-based SLAM problems irrespective of the front-end data
association method, and (iii) it generalizes existing approaches where data
association is assumed to be known or partially known.
We approach the problem by splitting it into an inner problem of estimating the
trajectory, landmark positions and data associations and an outer problem of
estimating the number of landmarks.  
Our approach 
creates useful and novel connections with existing techniques from
discrete-continuous optimization (\eg k-means clustering), which has the
potential to trigger novel research.  We demonstrate the proposed approaches in
extensive simulations and on real datasets and show that the proposed techniques
outperform typical data association baselines and are even competitive against
an  ``oracle'' baseline which has access to the number of landmarks and an
initial guess for each landmark.
\end{abstract}

%% file: sections/intro.tex


\section{Introduction}
\label{sec:intro}

This paper studies landmark-based Simultaneous Localization and Mapping (SLAM) and investigates the challenging case 
where data associations between measurements and landmarks are not given.
Consider a robot moving in an unknown environment and trying to estimate its own trajectory 
using odometry and detections of unknown landmarks. 
When data associations are given, this leads to a well-studied and now easy-to-solve setup, \ie
standard landmark-based SLAM~\cite{Cadena16tro-SLAMsurvey}. 
However, the problem becomes extremely challenging when data associations are unknown, \ie 
when the robot does not have prior knowledge about data association and does not even know how many unique
landmarks it has observed.
This case arises in practice when the detections do not provide rich information (\eg appearance, semantics) to perform data association, or after failures resulting from incorrect data associations (where one needs to recover an estimate by reasoning over potentially wrong associations).

The problem of data association in SLAM has been extensively studied~\cite{bar1990tracking}, see Section~\ref{sec:relatedWork} for 
a review.
However, most papers consider an \emph{incremental} setup where data associations are greedily established at each 
time step (\eg one has to associate newly acquired measurements to the existing set of landmarks). The setup where all associations have to be established at the same time (\ie a \emph{batch} optimization setup) 
has not received the same amount of attention. The batch setup is relevant for several reasons.
First, the incremental setup makes hard data-association decisions at each step, but data associations that look plausible at a given time may turn out to be wrong after more measurements are collected; on the other hand, a batch setup simultaneously reasons over the entire history of measurements and looks for an optimal association that is consistent across all measurements.
Second, typical incremental data association techniques are based on probabilistic tests that rely on
the measurement covariances; however, the covariances are often hand-tuned and may not be reliable, hence calling for an alternative framework that is less sensitive to the covariance tuning.
Finally, incremental data association may fail after an incorrect data association (or may accumulate large drift if correct data associations are discarded); therefore, it is desirable to have a batch approach that revisits all data associations and corrects mistakes, in order to recover after SLAM failures.


\myParagraph{Contribution} This paper investigates batch-SLAM with unknown data associations. 
Our first contribution (Section~\ref{sec:problemFormulation}) is to provide a formulation for the 
inner problem, which simultaneously optimizes for the robot poses, landmark positions, and data associations assuming the number of landmarks is given. We then present a formulation to estimate the number of landmarks.

Our second contribution (Section~\ref{sec:solvers}) presents algorithms to solve the optimization problems arising in our formulations. Our algorithm for the inner problem is based on an alternation scheme, which alternates the estimation of the robot poses
 and landmark positions with the choice of data associations. The algorithm is not guaranteed to compute optimal solutions due to the hardness and combinatorial nature of the optimization problems it attempts to solve, but it works well in practice.

 Finally, we demonstrate the proposed approaches in extensive simulations and on real datasets (Section~\ref{sec:experiments}), and show that 
the proposed techniques outperform typical data association baselines and are even competitive against an  ``oracle'' baseline which has access to the number of landmarks and an initial guess for each landmark.

%% file: sections/relatedWorks.tex

\section{Related Work}
\label{sec:relatedWork}
The data association problem of associating measurements
to states of interest has been extensively studied in the target tracking
literature \cite{bar1990tracking} and later computer vision and robotics. Data associations can be made in the front-end via feature descriptors \cite{mur2017orb} and recently via deep learning \cite{li2021odam}. However, these front-end methods are typically sensor-dependent and require additional information (\eg landmark appearance) from the sensor other than pure positional measurements of the landmarks. We are interested in the case where such information is not available and data associations must be inferred from the positional measurements. Standard robust optimization methods based on M-estimators used in SLAM (\eg \cite{Sunderhauf12iros,Agarwal13icra,Yang20ral-GNC}) are not directly applicable to such scenarios in an efficient manner (\ie they would require adding robust losses for all possible associations). Therefore, in the following we focus our attention on techniques designed for data association in landmark-based SLAM.

 Maximum likelihood data association 
 \cite{bar1990tracking} is a classical approach where one first rejects unlikely associations for which the
 Mahalanobis distance between measurements and their predictions is greater than
 a statistical threshold. This results in a number of
 \emph{individually compatible} association candidates for each measurement,
 among which one picks the one with the maximum likelihood. 
 In its most basic form, this approach processes measurements in a sequential
 manner and makes a decision for each measurement independently.
 Alternatively, one can solve a linear assignment problem between
 a batch of measurements and landmarks in which the cost of each measurement-landmark
 assignment is set to negative log-likelihood if they are \emph{individually
 compatible}, and to infinity otherwise \cite{bar1990tracking}.

 Joint compatibility branch and bound (JCBB) \cite{neira2001data}
 is a popular technique for solving the data association problem
 \emph{simultaneously} for a batch of measurements. This method leverages
 correlations between the estimated positions of landmarks and seeks the largest
 set of \emph{jointly compatible} associations. JCBB uses the branch and bound
 technique to prune the exponentially large search space of the problem.

The combined constraint data association (CCDA) method \cite{bailey02thesis} forms the \emph{correspondence graph} whose nodes correspond to individually compatible associations and undirected edges correspond to a pair of mutually compatible associations. CCDA
then aims to find the largest set of individually
compatible associations that are also \emph{pairwise} compatible by
equivalently searching for maximum clique in the correspondence graph.
A similar idea is proposed in \cite{Lusk21icra-clipper} in which the authors propose to form an
edge-weighted version of the correspondence graph and search for  
the densest clique.


Another approach is CLEAR \cite{fathian2020clear}, which identifies landmarks by clustering raw measurements using a
spectral graph clustering-type method. This is done by forming an association graph whose
nodes correspond to measurements and whose edges specify potential matches based
on a compatibility criterion. Then, CLEAR uses spectral graph-theoretic methods to
first estimate the number of distinct landmarks $K$, and then to rectify the 
association graph to obtain a disjoint union of $K$ cliques, each of which corresponds to a unique landmark.

Unlike our approach, the abovementioned works aim to find an explicit
set of associations \emph{prior} to estimating  robot trajectory
and the map in each time step. An alternative approach is proposed in
\cite{Dellaert00cvpr-correspondenceFreeSfm} where the authors treat associations
as latent variables and propose a scheme based on expectation-maximization (EM)
for estimating poses and 3D points. This is done by maximizing the
expected value of the log-likelihood function where the expectation is taken
over all possible associations. Similar approaches are proposed in \cite{Bowman17icra,michael2022probabilistic} where,
in addition to geometric information, semantic information is also incorporated.
Unlike \cite{Dellaert00cvpr-correspondenceFreeSfm,Bowman17icra,michael2022probabilistic}, in this
 work we aim to find the \emph{most likely}
 estimate for robot poses, landmark positions, and \emph{associations}, as well as the number of landmarks given
 the collected measurements. 

In a separate line of work \cite{mullane2011random}, the map is modeled as a random finite set (a
finite-set-valued random variable), inferring the number of landmarks from the collected measurements and circumventing the need for conventional data
association.

Our formulation for the inner problem of simultaneously estimating the robot poses, landmark positions and data associations has a strong connection to the max-mixture formulation \cite{Olson12rss} in the sense that landmark choices for a measurement can be modeled as different modes in the mixture. A max-mixture-type method for data association for semantic SLAM is proposed in \cite{doherty2020probabilistic} where additional semantic label and confidence information of landmarks is available to assist data association. Our inner problem is a special case of the general formulation recently presented in the discrete-continuous smoothing and mapping \cite{doherty2022discrete}. In Section~\ref{sec:experiments}, we compare our method with the alternating minimization approach proposed in \cite{doherty2022discrete} (see Oracle baseline).

%% file: sections/problemFormulations.tex


\section{Problem Formulations: \ProbNameLong (\probName)}
\label{sec:problemFormulation}

We break the \ProbNameLong (\probName) problem into an inner problem and an
outer problem. The inner problem solves for the robot poses, landmark positions
and data associations assuming the number of landmarks is known. The outer
problem estimates the number of landmarks assuming one can solve the inner problem. We first introduce our notations.
Let us denote the robot trajectory as $\vxx = \{\vxx_1, \ldots, \vxx_n\} \in \SE{d}^n$, 
where $\vxx_i \in \SE{d}$ is the (unknown) $i$-th pose of the robot and  
$\SE{d}$ is the group of poses in dimension $d=2$ or $3$;
in particular, $\vxx_i = (\MR_i,\vt_i)$, and $\MR_i$ and $\vt_i$ are the  rotation and translation components of the pose. 
We denote the (unknown) position of the $j$-th landmark as $\vy_j \in \Real{d}$; we then store all landmark position estimates as columns of a matrix $\vy$, the \emph{landmark matrix}. Finally, we denote the (given) measurement of a landmark position as $\vzbar_k \in \Real{d}$, for $k \in [m] \triangleq \{1,\ldots,m\}$, 
and use the standard generative model for the measurements:
\bea
\label{eq:ldmkmeasurement}
\vzbar_k = \MR_{i_k}\tran(\vy_{j_k}  - \vt_{i_k}) + \vepsilon_k
\eea
where a landmark $\vy_{j_k}$ is observed in the local frame of robot pose $i_k$,
up to zero-mean Gaussian measurement noise $\vepsilon_k$. 
Since we know the timestamp at which a measurement is taken, we can associate the measurement with the corresponding robot pose index $i_k$ (\eg the $k$-th measurement taken from the second robot pose has $i_k=2$). However we do not know 
what is the index $j_k$ of the landmark measured by $\vzbar_k$ (\ie we have unknown data associations). 
With this notation, we can now introduce our problem formulations.

\subsection{\probName Inner Problem}

We provide a problem formulation for the inner problem of \probName to estimate the robot poses and landmark positions as well as the data associations given the number of landmarks.

\begin{problem}[\probName inner problem]\label{prob:inner}
\bea
\label{eq:inner}
\min_{ \substack{\vxx \in \SE{d}^n \\ \vy \in \Real{d \times K} } } & 
 f_{\text{\normalfont odom}} (\vxx) +  \!\displaystyle \sum_{k=1}^m \min_{j_k \in [K]} \| \MR_{i_k}\tran (\vy_{j_k}  \!-\! \vt_{i_k}) \!-\! \vzbar_k \|_{\Sigma}^2 
\eea
where $K$ is the number of landmarks (\ie the number of columns of the matrix $\vy$).
The first term $f_{\text{\normalfont odom}} (\vxx)$ in~\eqref{eq:inner} 
 is the standard odometric cost (which is only a function of the robot trajectory)\footnote{
We refer the reader to standard references~\cite{Cadena16tro-SLAMsurvey,Rosen18ijrr-sesync} for the expression of $f_{\text{odom}}$, which is irrelevant for the 
rest of the discussion in this paper.}; the second term is similar to the one arising in landmark-based SLAM, but also
includes an optimization over the unknown data association $j_k$. Each summand in this term can be viewed as a max-mixture by considering each association choice as a mode in the mixture \cite{Olson12rss}.
\end{problem}

\begin{proposition}[Monotonicity\footnote{\label{notesuppl}The proofs are provided in the \supplementary.}]\label{prop:monotonicity}
Let us define the function $f_{\text{\normalfont slam}}(\vxx,\vy,K)  =
f_{\text{odom}} (\vxx) +  \!\textstyle \sum_{k=1}^m \min_{j_k \in [K]} \|
\MR_{i_k}\tran (\vy_{j_k}  \!-\! \vt_{i_k}) \!-\! \vzbar_k \|_{\Sigma}^2$, which
is the objective function in~\eqref{eq:inner}. After minimizing over $\vxx$ and
$\vy$, it remains a function of $K$, namely $f_{\text{\normalfont slam}}^\star(K)$. 
The function $f_{\text{\normalfont slam}}^\star(K)$ is monotonically
decreasing in $K$ and attains a minimum $f_{\text{\normalfont slam}}^\star(K) = 0$ when $K=m$. 
\end{proposition}

\begin{remark}[Hardness] \label{remark:hard}
 Problem~\ref{prob:inner} is NP-hard even when the optimal poses are known: in
 the next section we show that it indeed is equivalent to $K$-means clustering when
 the optimal poses and the number of landmarks are known, which is known to be NP-hard (see Proposition~\ref{prop:clustering}). 
\end{remark}

\subsection{\probName Outer Problem}

We now present our formulation to the outer problem of estimating the number of landmarks $K$ in Problem~\ref{prob:betaform}.

\begin{problem}[\probName as simplest explanation]\label{prob:betaform} Using the definitions in Proposition \ref{prop:monotonicity}, we write an outer minimization over K to estimate the number of landmarks ($K$):
\bea
\label{eq:formulationLmkCount}
\min_{K \in \Natural{}} &
\overbrace{\left( \displaystyle\min_{ \substack{\vxx \in \SE{d}^n, \vy \in \Real{d \times K}  } }  
 f_{\text{\normalfont slam}} (\vxx,\vy, K)  \right)}^{f_{\text{\normalfont slam}}^\star(K)}
  + \beta K  
\eea
\end{problem}

Note that $f_{\text{\normalfont slam}}^\star(K)$ decreases as we allocate more landmarks (\ie as $K$ increases) according to Proposition~\ref{prop:monotonicity}. On the contrary, $\beta K$ increases as more landmarks are allocated. The formulation in~\eqref{eq:formulationLmkCount} thus minimizes a cost that trades-off the SLAM error with the cost $\beta$ ($\beta > 0$) of using more landmarks. It essentially follows Occam's razor: we try to explain the measurements with the minimal number of landmarks possible. Problem~\ref{prob:betaform} together with Problem~\ref{prob:inner} simultaneously decides the SLAM estimates $\vxx$ and $\vy$,  while also deciding the number of landmarks (\ie the number of columns in $\vy$) and the data associations (minimization over $j_k$ in~\eqref{eq:inner}). The following remarks give some insights on the shape of the cost function in~\eqref{eq:formulationLmkCount} and on how to tune $\beta$.

\begin{remark}[Shape of ``U'']\label{remark:U}
Proposition~\ref{prop:monotonicity} states that the objective function
in~\eqref{eq:formulationLmkCount} is the sum of a monotonically decreasing
function of $K$ (\ie $f_{\text{\normalfont slam}}^\star(K)$) and a monotonically
increasing linear function of $K$ (\ie $\beta K$). This suggests that for sufficiently large values of $\beta$, the overall cost function $f(K) = f_{\text{\normalfont
slam}}^\star(K) + \beta K$, roughly has a ``U'' shape,
which starts at a large value for $K=1$ (this is the case where all measurements
are associated to a single landmark, in which case the cost
$f_{\text{\normalfont slam}}^\star(K)$ is typically large), then reaches a minimum for a suitable value of $K$, and then increases again until reaching a value $f(m) = \beta m$ for $K = m$. This insight will be useful to design our solver in~Section~\ref{sec:solver-km}, whose outer loop searches over potential values of $K$.
\end{remark}

\begin{remark}[Interpretation of  $\beta$] \label{remark:setBeta}
The parameter $\beta$ describes the cost of using one more landmark to explain the measurements. Choosing a large $\beta$ leads to underestimating the number of landmarks, while a small $\beta$ leads to overestimating it. 
More specifically, consider having a set of measurements $\calM_j$ for landmark $j$. The corresponding residual errors for these measurements will be $r_j \doteq \sum_{k \in \calM_j} \| \MR_{i_k}\tran (\vy_{j}  - \vt_{i_k}) - \vzbar_k \|_{\Sigma}^2$. 
Intuitively, when $r_j$ exceeds $\beta$, the optimization may prefer ``breaking'' the measurements across two or more landmarks (\ie increasing $K$). We explain how we set $\beta$ in Section~\ref{sec:baselines}.

\end{remark}

%% file: sections/solvers.tex
\section{Algorithms for \probName}
\label{sec:solvers}

\subsection{Algorithms for Problem~\ref{prob:inner}}
\label{sec:solver-km}


This section proposes a heuristic algorithm to solve Problem~\ref{prob:inner}.
Before introducing the algorithm, 
we provide fundamental insights that connect Problem~\ref{prob:inner} with Euclidean clustering and standard
landmark-based SLAM.  

\begin{proposition}[Clustering\textsuperscript{\ref{notesuppl}}]\label{prop:clustering}
Given $K$ and for fixed $\vxx$, Problem~\ref{prob:inner} becomes an Euclidean clustering problem assuming the measurement noise covariances $\Sigma$ are isotropic. In this clustering problem, the landmarks $\vy$ are the cluster centers and the data associations $j_k$ are equivalent to the cluster assignments.
\end{proposition}

\begin{proposition}[SLAM\textsuperscript{\ref{notesuppl}}]\label{prop:slam}
Given $K$ and the data associations $j_k$ for each measurement $\vzbar_k \in \Real{d}$, $k=1,\ldots,m$, 
Problem~\ref{prob:inner} reduces to standard landmark-based SLAM.
\end{proposition}

Propositions~\ref{prop:clustering}-\ref{prop:slam} suggest a simple alternation scheme, where we alternate 
solving for the data associations $j_k, \forall k=1,\ldots,m$ (given an initial guess for $\vxx$),
 with solving $\vxx,\vy$ (given the data associations). The former can be solved
 using standard clustering approaches (\eg Lloyd's algorithm for k-means clustering with k-means++ initialization~\cite{Arthur07-kmeanspp})   
 according to Proposition~\ref{prop:clustering}; the latter can use standard landmark-based SLAM solvers according to Proposition~\ref{prop:slam}. We name our algorithm \nameKmeans due to the combined use of k-means and SLAM solvers.

The pseudo-code of \nameKmeans (inner problem solver) is in Algorithm~\ref{algo:inner}.
Algorithm~\ref{algo:inner} performs a fixed number of iterations, where
each iteration alternates between two steps:
\begin{itemize} 
	\item {\bf Data association update} (lines~\ref{line:proj}-\ref{line:kmeans}) computes the data associations $j_k$ and landmark positions $\vy$ given $\vxx$; this reduces to clustering per Proposition~\ref{prop:clustering}, which we solve (locally) using k-means and k-means++~\cite{Arthur07-kmeanspp};
	\item  {\bf Variable update} (line~\ref{line:slam}) computes the robot poses $\vxx$ and landmark positions $\vy$ given the data associations $j_k$ and the result of $\vy$ from the first step as initial guess; this reduces to standard landmark-based SLAM per Proposition~\ref{prop:slam}, hence it can be readily solved with both local~\cite{Kuemmerle11icra,gtsam} and global solvers~\cite{Rosen18ijrr-sesync}.
\end{itemize}




\begin{algorithm}[h!]
		\small
	\caption{\nameKmeans inner solver.\label{algo:inner}}
	\SetAlgoLined
	\KwIn{Odometry and corresponding initial guess $\vxx$; landmark measurements $\vzbar_k$, $k=1,\ldots,m$; number of landmarks $K$; $\text{maxIterations}$}
	\KwOut{Estimate of robot poses $\vxx$; landmark positions $\vy$; SLAM objective value $f_{\text{slam}}$.}
    
    \For{ $\text{\normalfont iter} = 1,2, \ldots, \text{\normalfont maxIterations}$  \label{line:inner}}{
			\tcc{\footnotesize Project each measurement to the world frame based on current
			estimate of	$\vxx$} $\vzhat_k = \MR_{i_k}
			\vzbar_k \!+\! \vt_{i_k}$,
			\, \text{for $k \in [m]$} \label{line:proj}\;
    		$\text{initialize } \vy \text{ by k-means++}(\{\vzhat_1,\ldots,\vzhat_m\}, K)$\label{line:kmeans++}\;
    		$\{\vy, j_1,\ldots,j_m\} = \text{k-means}(
			\{\vzhat_1,\ldots,\vzhat_m\}, K)$\label{line:kmeans}\;
    		$\{\vxx,\vy\} = \text{SLAM}(\vxx,\vy,\{j_1,\ldots,j_m\})$\label{line:slam}\;
    	}
    


	\Return{$\{\vxx,\vy,f_{\text{\normalfont slam}}(\vxx,\vy,K)\}$.}
\end{algorithm}


\subsection{Algorithms for Problem~\ref{prob:betaform}}
\label{sec:solver-beta}

In consistency with Remark~\ref{remark:U}, we also empirically observe $f(K) = f_{\text{slam}}^\star(K) + \beta K$ has roughly a U-shape. Therefore, we simply use a gridding method to solve for $K^\star$. Naive gridding, which probes the objective function at every $K$ ($K = 1,\ldots,m$), requires $m \cdot \text{maxIterations}$ invocations of the landmark-based SLAM solver. For practical problems where the number of measurements $m$ is in the thousands and considering that in our experiments $\text{maxIterations} = 15$, this is too expensive. We thus introduce a multi-resolution gridding technique, whose pseudo-code is in Algorithm~\ref{algo:beta}.

\begin{algorithm}[h!]
		\small
	\caption{\nameKmeans solver.\label{algo:beta}}
	\KwIn{Odometry and corresponding initial guess $\vxx$; landmark measurements $\vzbar_k$, $k=1,\ldots,m$; $\text{maxIterations}$; parameter $\beta$; number of $K$'s to probe at each level $N_K$.}
	\KwOut{Estimate of robot poses $\vxx$; landmark number ($K$) and positions $\vy$.}
    $\{K_1, K_2, ..., K_{N_K}\} = \text{UniformDiv}([1,m], N_K)$\label{line:interval1m}\;
    \While{$|K_1 - K_2| > 1$ \label{line:1resol}} {
        \For{$K = K_1, K_2, ..., K_{N_K}$\label{line:forK}} {
            $\{\vxx\at{K}, \vy\at{K}, f_{\text{slam}}\at{K}\} = \text{Alg.}~\ref{algo:inner}(\vxx, \{\vzbar_1,\ldots,\vzbar_m\}, K, \text{maxIterations})$\;
            $f\at{K} = f_{\text{slam}}\at{K} + \beta K$\; \label{line:eval_f}
            \tcc{\footnotesize Save best result}
        	\If{$f\at{K} \leq f^\star$ \label{line:optCost}}{
        		$f^\star \leftarrow f\at{K}$\;
        		$\{\vxx^\star,\vy^\star\} \leftarrow  \{\vxx\at{K},\vy\at{K}\}$\;
        		$K_{n^\star} \leftarrow K$\;
        	}
        }
        $\{K_1, ..., K_{N_K}\} = \text{UniformDiv}([K_{n^\star-1},K_{n^\star+1}], N_K)$\label{line:shrinkint}\;
    }
	\Return{$\{\vxx^\star,\vy^\star,K_{n^\star}\}$.}
\end{algorithm}

 In Algorithm~\ref{algo:beta}, we iteratively grid the interval where the optimal $K$ could lie into smaller intervals until we find the exact value of $K$ in a resolution-one grid. Specifically, we start with the interval $[1, m]$ (line~\ref{line:interval1m}) and uniformly evaluate $f^\star_{\text{slam}}(K) + \beta K$ at $N_K$ values ($N_K$ is set to 11), $K_1, K_2, ..., K_{N_k}$, within this interval and including both end points (lines~\ref{line:forK}-\ref{line:eval_f}). We find the optimal $K_{n^\star}$ out of these $N_k$ values (line~\ref{line:optCost}), shrink the interval down to $[K_{n^\star-1}, K_{n^\star+1}]$ (line~\ref{line:shrinkint}), and repeat the process until the resolution is one (line~\ref{line:1resol}). This multi-resolution strategy makes approximately $N_k\log_{\frac{N_k}{2}}(m)$ evaluations of $f_{\text{slam}}^\star(K)$ compared to $m$ evaluations in the naive gridding. For $m=1000$ and $N_K=11$, this is 45 versus 1000 evaluations.

%% file: sections/experiments.tex
\section{Experiments}
\label{sec:experiments}

This section shows that the proposed \nameKmeans algorithm computes accurate estimates across multiple simulated and real datasets, and it is competitive against two baselines, one of which has access to the true number of landmarks.
We first introduce the experimental setup and datasets 
(Section~\ref{sec:datasets}) and the  baseline techniques (Section~\ref{sec:baselines}).
We then evaluate the techniques in terms of accuracy (Section~\ref{sec:accuracy}). An ablation study is provided in the \supplementary.


\subsection{Datasets and Setup}\label{sec:datasets}

\myParagraph{Datasets} We evaluate \nameKmeans along with baseline techniques on 
(i) two fully synthetic ``grid'' datasets (2D and 3D), 
(ii) two semi-real datasets (Intel 2D and Garage 3D) where the trajectories and odometry measurements are real, but the landmarks are synthetic, and 
(iii) a real dataset (Indoor). 
See the \supplementary~for visualizations of the datasets.

For the two synthetic datasets, we first generate a grid-like ground-truth trajectory and uniformly sample ground-truth landmark positions around the trajectory. We then add Gaussian noise to create odometry and landmark measurements following eq.~\eqref{eq:ldmkmeasurement}. Each landmark is observed by a fixed number of poses that are closest to it in the ground-truth positions. 

For the two semi-real datasets (Intel and Garage), we use two real datasets for pose graph optimization~\cite{kummerle2011g, g2odata}, which include both odometry measurements and loop closures. 
We obtain a proxy for the ground-truth trajectory by optimizing the original pose graph with loop closures.
Then we remove the loop closures from the dataset, and add landmarks following the same procedure used for the synthetic datasets.

The real dataset (Indoor) is a benchmarking dataset available online~\cite{mathworksdata}. 
In this dataset, a robot navigates in a real office environment and observes a number of AprilTags~\cite{april} as landmarks along the trajectory (we only use the ID of the AprilTags for evaluating different methods). 

\myParagraph{Setup} In our synthetic and semi-real experiments, we sweep through
different parameter values that define the map (\eg odometry noise level, number
of landmarks, and landmark measurement noise level). Twenty Monte Carlo simulations are performed for each set of parameters where we vary noise realizations and landmark placement. The default map parameter values can be found in Table~\ref{tab:simparams}.

\input{includes/table-params}


\subsection{Compared Techniques and Baselines}\label{sec:baselines}

\myParagraph{Proposed Technique} \nameKmeans uses the Levenberg-Marquardt optimizer in GTSAM~\cite{gtsam} 
for landmark-based SLAM. We set $\text{maxIterations}=15$. A rule of thumb is to set $\beta$ to the maximum expected residuals for the measurements of a single landmark. Assuming that a landmark $j$ is observed $n = |\calM_j|$ times and that the measurements are affected by Gaussian noise, $r_j$ (defined in Remark~\ref{remark:setBeta}) at the ground truth is distributed according to a $\chi^2$ distribution, hence
$\prob{r_j \leq \gamma^2} = p$, where $\gamma^2$ is the quantile of the $\chi^2(dn)$ distribution with 
$d \times n$ degrees of freedom and lower tail probability equal to $p$. 
As a heuristic, we set $\beta = \gamma^2$ where $\gamma$ can be computed for a given choice of tail probability (in our experiments $p = 0.997$) and for an average number of landmark observations $n$. Assuming having knowledge of $n$, the values of $\beta$ used are 41.72 (Grid 2D), 55.64 (Grid 3D), 68.94 (Intel), and 94.47 (Garage). In practice, a rough estimate of $K$ can be obtained by clustering the projected measurements (line~\ref{line:proj} in Algorithm~\ref{algo:inner}) and applying the elbow method \cite{thorndike1953belongs} to estimate the number of clusters (\ie $K$). An estimate of $n$ can be then derived from this estimate of $K$ given the total number of landmark measurements which is known. The accuracy of \nameKmeans is not sensitive to the exact value of $\beta$ as seen in our sensitivity analysis in Table~\ref{tab:realresults} and in the \supplementary. 


\myParagraph{\nameAlt Baseline} The \nameAlt baseline is an alternation scheme that has access to the true number of landmarks and an initial guess for each landmark.
The baseline alternates two steps. In the first step, 
for each measurement $\vzbar_k$  it finds the landmark having a position estimate
$\vy_j$ that minimizes $\| \MR_{i_k}\tran (\vy_{j}  \!-\! \vt_{i_k}) \!-\!
\vzbar_k \|$. After making associations for all the measurements, it solves the
SLAM problem with the estimated associations. This process repeats until
convergence or reaching a user-specified number of iterations. This method is similar to the technique recently proposed in \cite{doherty2022discrete}. Note that this strong baseline assumes extra knowledge (\ie number of landmarks, initial guess for the landmarks) that is not available to the other techniques and in real scenarios. The initial guess for the landmarks is computed by projecting the first measurement of each landmark (\ie measurement from the first pose that observes the landmark) into the world frame (in the same way as line~\ref{line:proj} in Algorithm~\ref{algo:inner}).

\myParagraph{\nameML Baseline} The second baseline is the popular Maximum Likelihood data association used in an
incremental formulation of SLAM, where odometry is propagated sequentially and newly acquired measurements are incrementally associated to the most likely landmark estimate; see~\cite{bar1990tracking}.
At each time step, for each measurement, it selects a set of admissible landmarks by performing a $\chi^2$-test between the landmark estimates $\vy_{j}$ and the measurement $\vz_{k}$. The measurement is associated to the most likely landmark, while each landmark is only associated to one of the newly acquired measurements. The measurements that are not associated to any landmark are initialized as new landmarks. After data associations are established, batch optimization with the Levenberg-Marquardt optimizer \cite{levenberg_1944,marquardt_1963} is performed to compute the SLAM estimate at each time step. Additional implementation details are in the \supplementary.

\myParagraph{\nameOdom Baseline} This baseline only computes an odometric estimate of the robot 
trajectory by concatenating odometry measurements without using landmark measurements. We report the accuracy of this baseline as a worst-case reference.

\vspace{-2mm}
\subsection{Results: Accuracy}\label{sec:accuracy}

\input{sections/fig-accuracy2}

We measure the accuracy of the techniques by computing the Absolute Trajectory Error (ATE) of the trajectory estimated by each technique with respect to an optimized trajectory obtained via standard landmark-based SLAM with ground-truth data associations. Therefore, if the algorithms estimate all data associations correctly, the ATE will be zero. Unlike the normal ATE with respect to the ground-truth trajectory, this metric aims to capture data association correctness. In the following, we evaluate the ATE for varying levels of odometry noise, number of landmarks, and landmark measurement noise. Then we provide summary statistics for the real dataset (Indoor) in Table~\ref{tab:realresults}.

\myParagraph{Influence of odometry noise} 
 Fig.~\ref{fig:odomNoise} shows the ATE for increasing odometry noise in our synthetic datasets. 
 For low and moderate odometry noise levels, \nameKmeans generally outperforms other techniques. When the odometry noise is high, \nameML starts to outperform \nameAlt and \nameKmeans. The reason is that the poor odometry causes the initial projected measurements of a landmark (line~\ref{line:proj} in Algorithm~\ref{algo:inner}) to be too far to cluster and associate. As an incremental method, \nameML reduces the odometry drift incrementally, alleviating this problem.
 Fig.~\ref{fig:nrLmksa} shows the estimated number of landmarks for \nameKmeans and \nameML (\nameAlt is given the ground-truth number). We observe that \nameML tends to largely overestimate the number of landmarks. 
 In contrast, \nameKmeans provides fairly accurate estimation of the number at low to medium odometry noise. It only overestimates the number of landmarks at high odometry noise. We note that underestimating the number of landmarks in SLAM can lead to catastrophic failures due to incorrect associations. Overestimation, however, generally has less severe consequences, causing the method to miss some of the correct associations and thus lose loop-closing information.
 
 

\myParagraph{Influence of number of landmarks and landmark noise} 
Fig.~\ref{fig:lmkNoiseAndNumber} shows the ATE across the four datasets for increasing 
landmark measurement noise (top row) and increasing number of landmarks shown as a percentage of the number of poses (bottom row). 
We observe that our \nameKmeans still outperforms all other techniques almost across the spectrum; it is only worse at high landmark noise where the raw odometry becomes the best among all the techniques (with the exception of the garage dataset). In this case, two factors come into play for \nameKmeans. First, as mentioned earlier, our approach can underestimate the number of landmarks under high noise (see Fig.~\ref{fig:nrLmksb}). Second, high landmark noise makes solving the inner problem (Problem~\ref{prob:inner}) more challenging in \nameKmeans by reducing the inter-cluster distances of the projected measurements. Solving the inner problem sub-optimally further affects the outer estimation of the number of landmarks, causing more performance loss. The Garage dataset is particularly challenging as its odometry translation covariance is specified to be large despite the odometry being accurate. The \nameML baseline is the most vulnerable to such mismatched covariances due to its reliance on them.
Number of landmarks is a relatively weak factor for data association. There is a slight trend of having less accurate data association as more landmarks are in the map, which is expected.



\input{includes/table-real}

\myParagraph{Performance on the real dataset}
Table~\ref{tab:realresults} reports the results on the real dataset. 
We observe that although being slow, \nameKmeans still dominates the others. Additionally, the performance is not sensitive to $\beta$. The highest accuracy is obtained over a wide range of values for $\beta\in[20,140]$. Discussions about the runtime are in the \supplementary. 




%% file: includes/table-params.tex

\begin{table}[t]
    \vspace{5pt}
    \smaller
    \centering
    \noindent
    \caption{
       Dataset default parameters. Values in bold are varied in our experiments. n/a stands for a dense covariance matrix not representable by a single value.
    }
    \begin{tabular}{c | c c c c c}
        \hline
        Dataset & 2D Grid & 3D Grid & Intel & Garage & Indoor \\
        \hline
        \# of poses & 500 & 216 & 942 & 1661 & 571 \\
        \# of landmarks & \textbf{100} & \textbf{43} & \textbf{94} & \textbf{166} & 14 \\
        \# of lm. measurements & 1000 & 430 & 1880 & 3320 & 233  \\
        odometry translation std. & \textbf{0.05} & \textbf{0.05} & 0.045 & 1 & 0.1 \\
        odometry rotation std. & \textbf{0.005} & \textbf{0.005} & 0.014 & n/a & 0.1 \\
        lm. measurement std. & \textbf{0.05} & \textbf{0.05} & \textbf{0.05} & \textbf{0.05} & 0.1 \\
        \hline
    \end{tabular}
    \label{tab:simparams}
    \vspace{-5pt}
\end{table}

%% file: sections/fig-accuracy2.tex
\begin{figure}[t]
    \centering
    \begin{subfigure}{0.5\linewidth}
        \centering
        \includegraphics[width=1.1\linewidth]{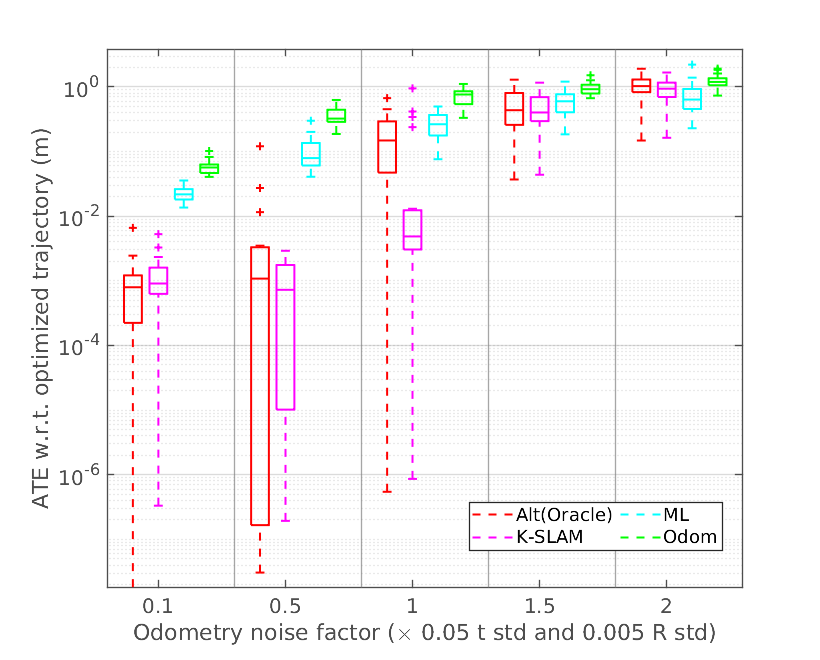}  
        \caption{Grid (2D)}
    \end{subfigure}%
    \begin{subfigure}{0.5\linewidth}
        \centering
        \includegraphics[width=1.1\linewidth]{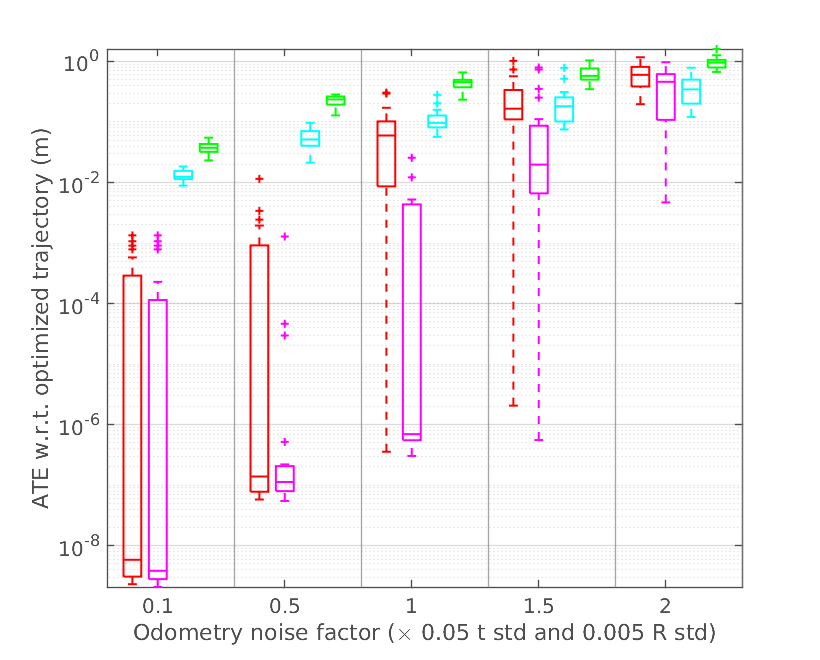}
        \caption{Grid (3D)\label{fig:odomNoise3D}}
    \end{subfigure}%
    \caption{Absolute trajectory error (ATE) for increasing odometry noise level on the synthetic grid datasets.\label{fig:odomNoise} \vspace{-3mm}}
\end{figure}

\begin{figure}[t]
    \centering
    \begin{subfigure}{0.5\linewidth}
        \centering
        \includegraphics[width=1.1\linewidth]{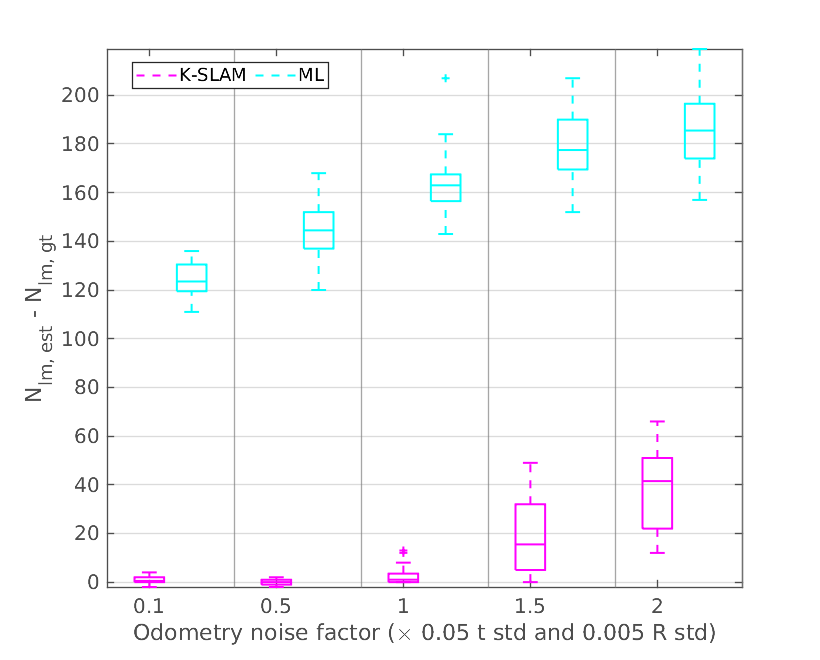}
        \caption{Odometry Noise\label{fig:nrLmksa}}
    \end{subfigure}%
    \begin{subfigure}{0.5\linewidth}
        \centering
        \includegraphics[width=1.1\linewidth]{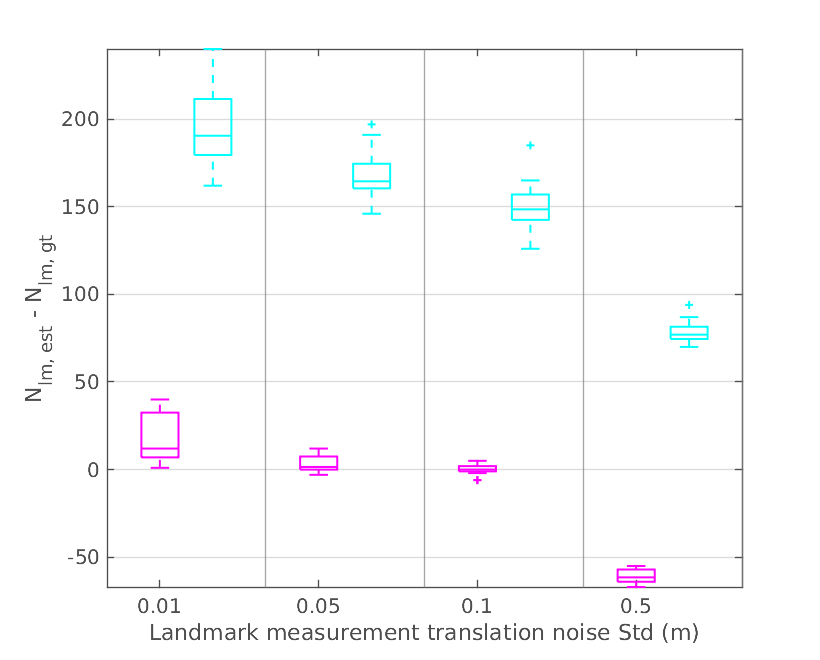}
        \caption{Landmark Noise\label{fig:nrLmksb}}
    \end{subfigure}%
    \caption{Difference between the estimated and the ground truth number of landmarks for \nameKmeans and the \nameML baseline on the 2D grid dataset with different noise levels.\label{fig:nrLmks}}
     \vspace{-6mm}
\end{figure}

\begin{figure*}[h]
    \centering
    \begin{subfigure}{0.25\linewidth}
        \centering
        \includegraphics[width=\linewidth]{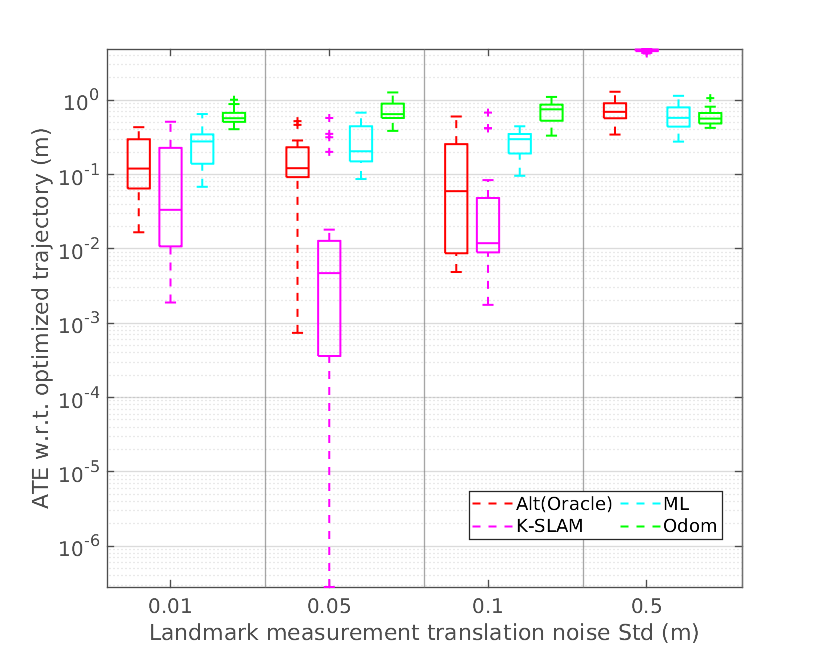}
    \end{subfigure}%
        \begin{subfigure}{0.25\linewidth}
        \centering
        \includegraphics[width=\linewidth]{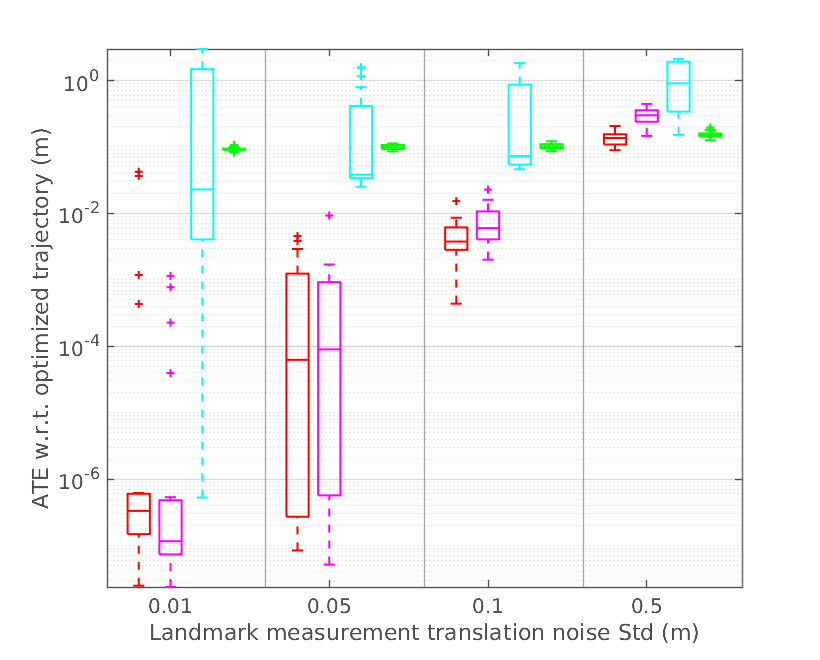}
    \end{subfigure}%
    \begin{subfigure}{0.25\linewidth}
        \centering
        \includegraphics[width=\linewidth]{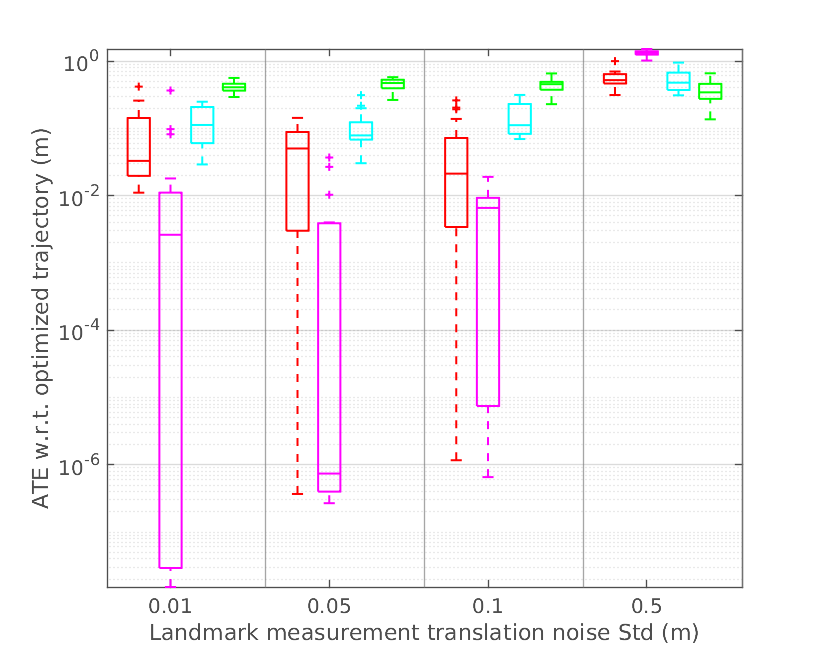}
    \end{subfigure}%
    \begin{subfigure}{0.25\linewidth}
        \centering
        \includegraphics[width=\linewidth]{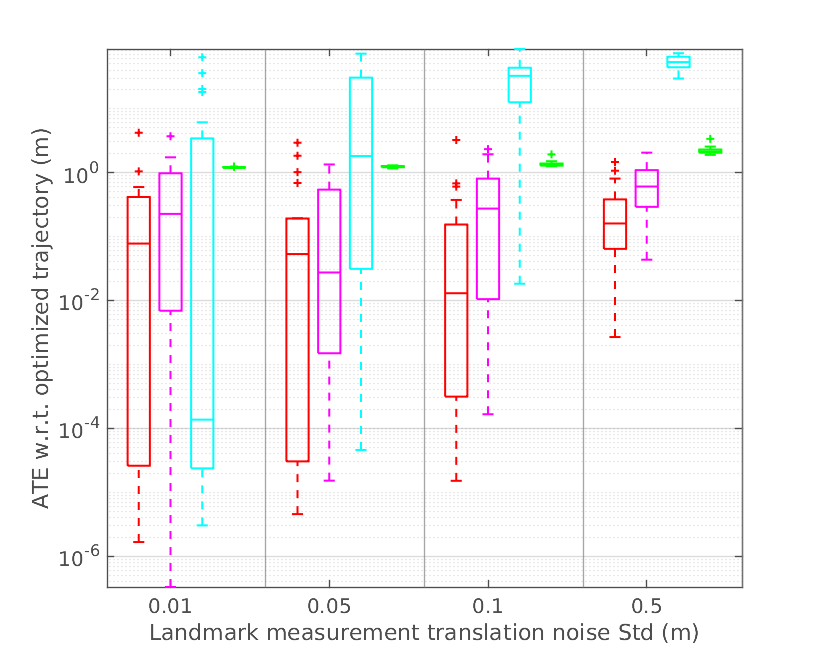}
    \end{subfigure}%
    \\
    \begin{subfigure}{0.25\linewidth}
        \centering
        \includegraphics[width=\linewidth]{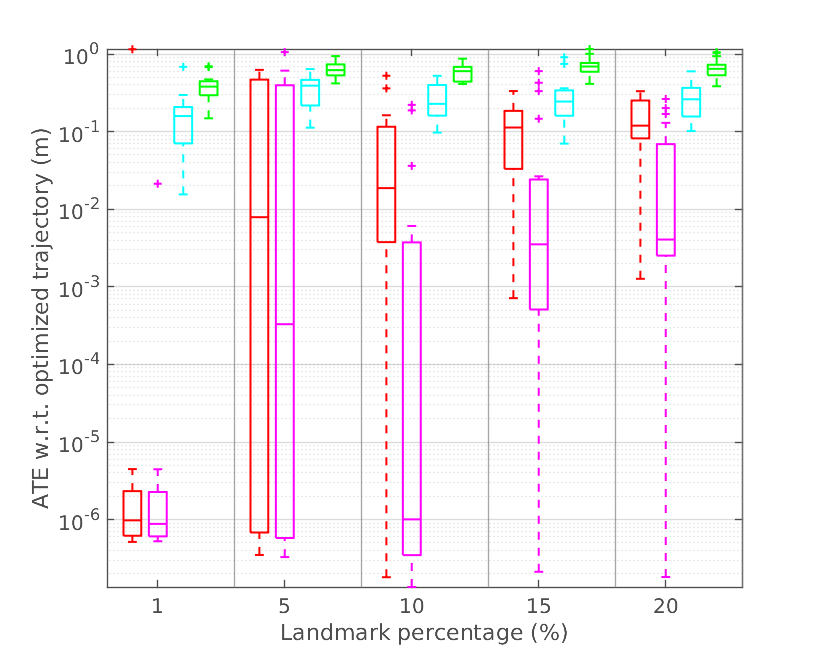}
        \caption{Grid (2D)}
    \end{subfigure}%
        \begin{subfigure}{0.25\linewidth}
        \centering
        \includegraphics[width=\linewidth]{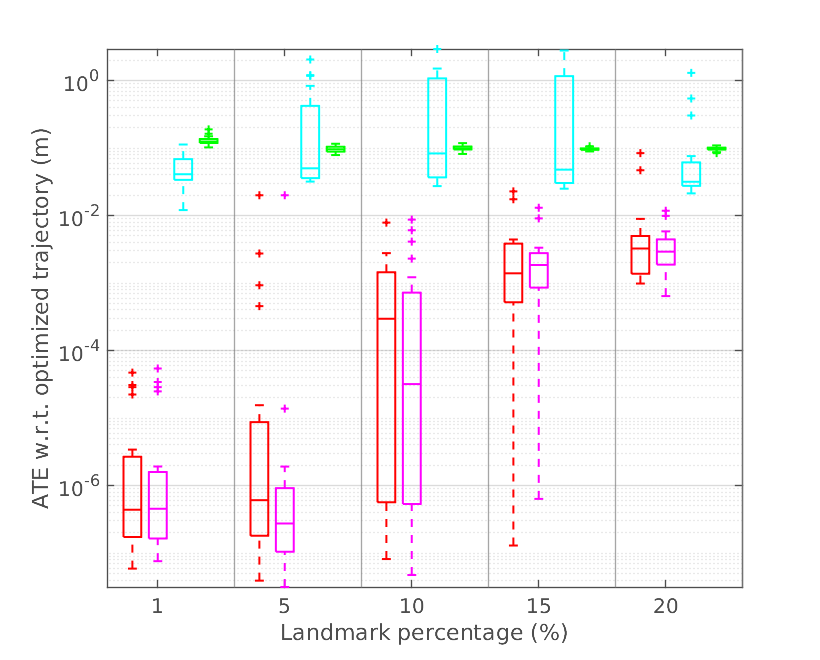}
        \caption{Intel (2D)}
    \end{subfigure}%
    \begin{subfigure}{0.25\linewidth}
        \centering
        \includegraphics[width=\linewidth]{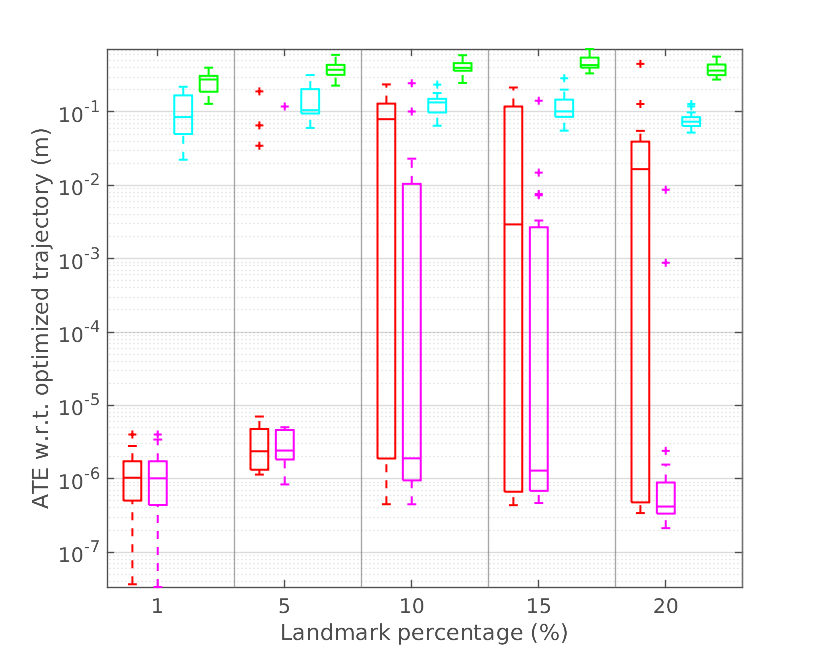}
        \caption{Grid (3D)\label{fig:lmkNoiseAndNumber3D}}
    \end{subfigure}%
    \begin{subfigure}{0.25\linewidth}
        \centering
        \includegraphics[width=\linewidth]{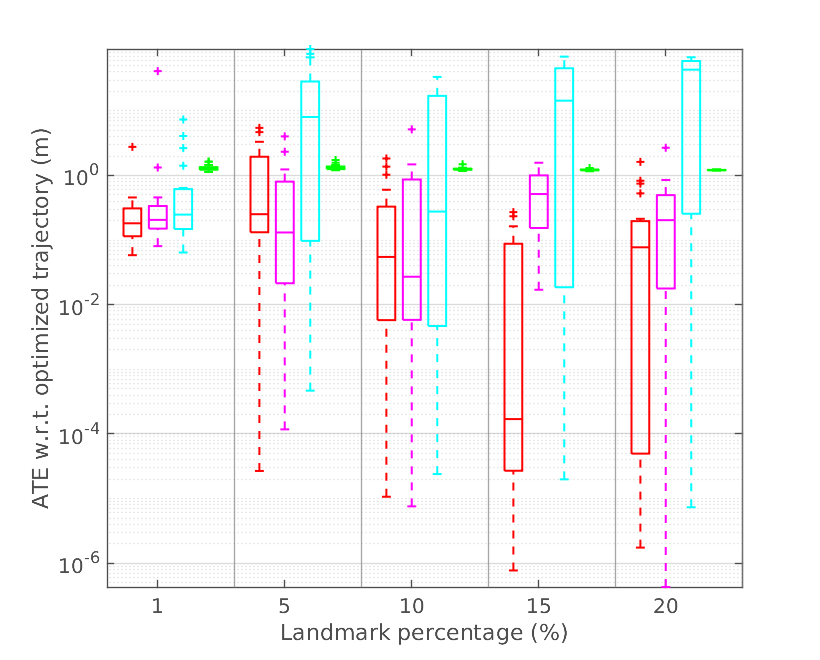}
        \caption{Garage (3D)}
    \end{subfigure}%
    \caption{Absolute trajectory error (ATE) for increasing landmark measurement noise and increasing number of landmarks (shown as a percentage of the number of poses).\label{fig:lmkNoiseAndNumber}\vspace{-3mm}}
    \vspace{-1mm}
\end{figure*}

%% file: includes/table-real.tex


\setlength{\tabcolsep}{4pt}

\begin{table}[tb]
    \footnotesize
    \centering
    \noindent
    \caption{
        ATE (m) and runtime (sec) on the Indoor dataset.
    }
    \begin{tabular}{c|cccccc}
        \hline
        \multirow{2}{3em}{Metric} & \multirow{2}{3em}{\nameOdom} & \multirow{2}{2em}{\nameML} & \multirow{2}{3em}{\nameAlt} & \multirow{2}{4em}{\nameKmeans $\beta=10$} & \multirow{2}{4em}{\nameKmeans $\beta=80$}  & \multirow{2}{4em}{\nameKmeans $\beta=240$}\\
        \\
        \hline
        ATE & 0.415  & 1.34 & 0.076 & 0.054 & \textbf{0.046} & 0.148 \\
        Runtime & n/a & 7.57 & 1.25 & 18.80 & 18.75 & 18.79 \\
        \hline
    \end{tabular}
    \label{tab:realresults}
    \vspace{-0.5cm}
\end{table}

%% file: sections/conclusion.tex
\section{Conclusion}
\label{sec:conclusion}

In this paper we investigated a challenging batch SLAM problem where the goal is to simultaneously estimate
the poses of the robot, the locations of unknown landmarks, and the data associations between measurements
and landmarks. We proposed problem formulations and algorithms for the resulting data-association-free SLAM problem. The algorithms were tested on a mix of synthetic and real datasets and perform better or on par with strong baselines that have access to privileged information. While these initial findings are encouraging, our algorithm does show limitations such as degraded performance at high noise and slow runtime. It would be desirable to better understand the behavior of $f_{\text{\normalfont slam}}^\star(K)$ for $K$ estimation and improve the robustness of association and clustering at high noise.

%% file: sections/appendix.tex

\appendix 
\section*{Proof of Proposition~\ref{prop:monotonicity}}
\begin{proof}
We need to show that 
$f_{\text{slam}}^\star(K-1) \geq f_{\text{slam}}^\star(K)$, for any $K \in \{1,2,\ldots,m\}$. 
Note that $f_{\text{slam}}^\star(K-1)$ and $f_{\text{slam}}^\star(K)$ are both optimization problems and the 
only difference is that the former imposes an extra constraint over the $\vy$ (in particular, that one more landmark has to match one of the existing landmarks). Therefore the feasible set of $f_{\text{slam}}^\star(K-1)$ is contained in the feasible set of $f_{\text{slam}}^\star(K)$, and thus $f_{\text{slam}}^\star(K-1) \geq f_{\text{slam}}^\star(K)$.
\end{proof}

\section*{Proof of Proposition~\ref{prop:clustering}}
\begin{proof}
$K$-means clustering   
requires partitioning $m$ points $\vzhat_k$, $k=1,\ldots,m$,  
into $K$ clusters such that the sum-of-squared-distances between points in a cluster and the cluster's centroid are minimized. 
More formally, a standard $K$-means clustering problem can be written as the following optimization problem:
\bea
\label{eq:clustering}
\min_{\vy \in \Real{d\times K}}  \sum_{k=1}^m \min_{j_k} \| \vy_{j_k} - \vzhat_k \|^2 
\eea
The problem essentially looks for $K$ cluster centroids $\vy_{j_k}$ (each one is a column of the matrix $\vy$)
while associating each point $\vzhat_k$ to a cluster centroid via $j_k$.

We now show that when $\vxx$ and $K$ are given, then Problem~\ref{prob:inner} can be written in the form~\eqref{eq:clustering}. When $\vxx$ and $K$ are given, Problem~\ref{prob:inner} becomes:
\bea
\label{eq:clustering2}
\min_{ \vy \in \Real{d \times K} } & 
 \!\displaystyle \sum_{k=1}^m \min_{j_k} \| \MR_{i_k}\tran (\vy_{j_k}  \!-\! \vt_{i_k}) \!-\! \vzbar_k \|^2 + \text{const.}
\eea
where we noticed that the first and last term in Problem~\ref{prob:inner} are constant (for given $\vxx$ and $K$) and are hence irrelevant for the optimization. 
Now recalling that the $\ell_2$ norm is invariant to rotation (and dropping constant terms), eq.~\eqref{eq:clustering2} becomes:
\bea
\label{eq:clustering3}
\min_{ \vy \in \Real{d \times K} } & 
 \!\displaystyle \sum_{k=1}^m \min_{j_k} \| \vy_{j_k}  \!-\! (\MR_{i_k}\vzbar_k \!+\! \vt_{i_k}) \|^2 
\eea
Now recalling that the poses (hence $\MR_{i_k}$ and $\vt_{i_k}$) are known, and
defining vectors $\vzhat_k \doteq \MR_{i_k}\vzbar_k \!+\! \vt_{i_k}$ we can
readily see eq.~\eqref{eq:clustering3} becomes identical to
eq.~\eqref{eq:clustering}, proving our claim.
\end{proof}

\section*{Proof of Proposition~\ref{prop:slam}}
\begin{proof}
The proof trivially follows from the fact that for given $K$ and data associations $j_k$, 
Problem~\ref{prob:inner} simplifies to (after dropping the constant term $\beta K$, which is irrelevant for the optimization once $K$ is given):
\bea
\label{eq:stdslam}
\min_{ \substack{\vxx \in \SE{d}^n \\ \vy \in \Real{d \times K} } } & 
 f_{\text{odom}} (\vxx) +  \!\displaystyle \sum_{k=1}^m \| \MR_{i_k}\tran (\vy_{j_k}  \!-\! \vt_{i_k}) \!-\! \vzbar_k \|^2 \nonumber 
\eea
which matches the standard formulation of landmark-based SLAM, and can be readily solved with both local~\cite{Kuemmerle11icra,gtsam} and global solvers~\cite{Rosen18ijrr-sesync}.
\end{proof}

\section*{Description of the ML Baseline} \label{sec:mldetails}
The incremental maximum likelihood approach includes two steps at each timestamp. The first step, for each measurement, selects a set of admissible landmarks by performing a $\chi^2$-test dependent on the {Mahalanobis distance} between the landmark estimate $\vy_{j}$ and the measurement $\vz_{k}$. We use a probability tail of $0.8$ for the test in all datasets except for Garage, where $0.9$ is used.
The second step associates the measurement with the most likely landmark (\ie the one achieving the {lowest negative loglikelihood score}), while ensuring that each landmark is only associated to one of the newly acquired measurements. This is done by forming a cost matrix with admissible landmarks and corresponding measurement candidates and solving for the optimal hypothesis with the Hungarian algorithm \cite{kuhn_1955,munkres_1957}.
The measurements that are not associated to a landmark are initialized as new landmarks using their projections in the world frame as the initial estimates.
At each timestamp, after data associations are established as described above, the \nameML baseline
 uses batch optimization with the Levenberg-Marquardt optimizer \cite{levenberg_1944,marquardt_1963} to compute the SLAM estimate. 

\section*{Visualizations of the Datasets} \label{sec:vizdatasets}
We visualize the five datasets in their default setup (Table~\ref{tab:simparams}) in Fig.~\ref{fig:mapviz}.
\input{sections/fig-datasets}

\section*{Sensitivity Analysis of $\beta$} \label{sec:sabeta}
\input{sections/fig-sensitivityBeta.tex}
We vary the value of $\beta$ in \nameKmeans and compare how the estimation accuracy changes with $\beta$ under different noise conditions in Fig.~\ref{fig:sensitivity}. Overall, the performance of \nameKmeans is not sensitive to the choice of $\beta$. In the default noise case, a range of $\beta$ values has almost the same accuracy. Under high odometry noise, \nameKmeans shows overestimation of the number of landmarks as seen in Section~\ref{sec:accuracy}. The overestimation is reduced as $\beta$ increases following our expectation from formulation~\eqref{eq:formulationLmkCount}. On the contrary, under the high landmark noise condition, underestimation of the number of landmarks is present, but it is alleviated as $\beta$ decreases. Even in the challenging high noise conditions, the ATE of \nameKmeans does not vary significantly across different values of $\beta$.

\section*{Ablation} \label{sec:ablation}
We present an ablation study on the Grid (3D) dataset in Fig.~\ref{fig:ablation} to understand the contributions of different components in our algorithm. The following observations are made in Fig.~\ref{fig:ablationa}. (i) We would like to know whether the provided initial guess actually helps \nameAlt. Comparing Alt (\nameAlt) to Alt (kmpp init) and \nameKmeans (given init + carry) to \nameKmeans (carry), we confirm that providing the initial guess for landmarks indeed helps improve the accuracy by a large margin (except for Alt (\nameAlt) at the highest noise level). (ii) We are also interested in whether clustering (line~\ref{line:kmeans} in Algorithm~\ref{algo:inner}) is better than the one-step association in the \nameAlt baseline. Comparing Alt (\nameAlt) and \nameKmeans (given init + carry), we observe a small improvement by doing iterative clustering for association. (iii) The largest improvement that helps \nameKmeans beat the \nameAlt baseline is line~\ref{line:kmeans++} in Algorithm~\ref{algo:inner} as seen from the performance gap between \nameKmeans (known $K$) and \nameKmeans (carry). The randomness from k-means++ greatly improves the accuracy (with the exception at the highest landmark noise level) by jumping out of local minima. Under high noise, the randomness becomes detrimental because it hampers convergence in the highly nonlinear cost function topology. (iv) We further observe that knowing the true $K$ does not necessarily lead to the best accuracy (see \nameKmeans versus \nameKmeans (known K)). The higher performance of \nameKmeans at lower noise levels is indeed caused by slight overestimation of the number of landmarks since overestimation can reduce the chance of map collapsing which may happen when there are fewer estimated landmarks than actual.

For runtime in Fig.~\ref{fig:ablationb}, the three variants of Algorithm~\ref{algo:inner} (given the ground-truth K) are faster than the \nameAlt baseline as they leverage efficient and mature k-means and k-means++ implementation. Therefore, the clustering steps (lines~\ref{line:kmeans++}-\ref{line:kmeans} in Algorithm~\ref{algo:inner}) bring better accuracy not at the expense of speed. The burden of estimating $K$ in an outer iteration scheme significantly increases the runtime for \nameKmeans.
\input{sections/fig-ablation}

\section*{Runtime Comparison} \label{sec:runtime}
The slow runtime of \nameKmeans in Table~\ref{tab:realresults} is attributed to the outer estimation of $K$ with a gridding approach as shown in the ablation study. \nameAlt is given the true $K$ so there is no outer estimation. \nameML only makes a greedy choice of whether to initialize a new landmark at every step so there is no significant burden from estimating the number of landmarks ($K$) and its runtime is also better than \nameKmeans. If a good initial guess of $K$ is available, we can possibly grid in a smaller interval and achieve better runtime for \nameKmeans.

%% file: sections/fig-datasets.tex

\begin{figure}[htb!]
    \centering
    \begin{subfigure}{0.5\linewidth}
        \centering
        \includegraphics[width=\linewidth]{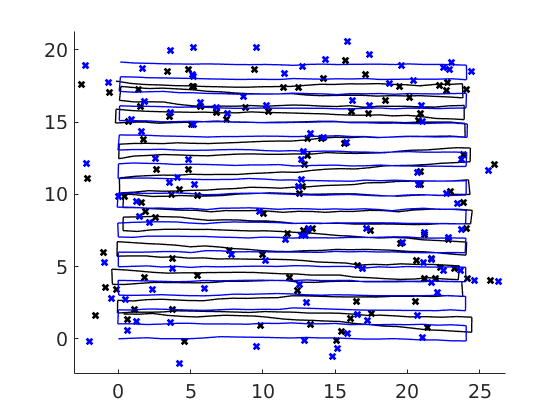}
        \caption{Grid (2D)}
    \end{subfigure}%
    \begin{subfigure}{0.5\linewidth}
        \centering
        \includegraphics[width=\linewidth]{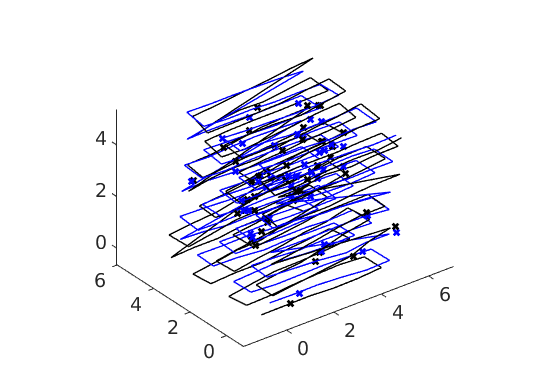}
        \caption{Grid (3D)}
    \end{subfigure}%
    \newline
    \begin{subfigure}{0.5\linewidth}
        \centering
        \includegraphics[width=\linewidth]{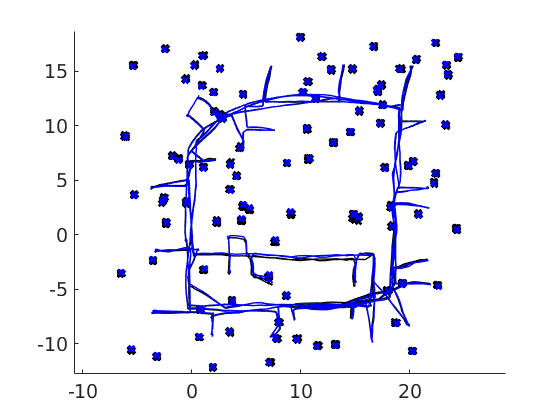}
        \caption{Intel (2D)}
    \end{subfigure}%
    \begin{subfigure}{0.5\linewidth}
        \centering
        \includegraphics[width=\linewidth]{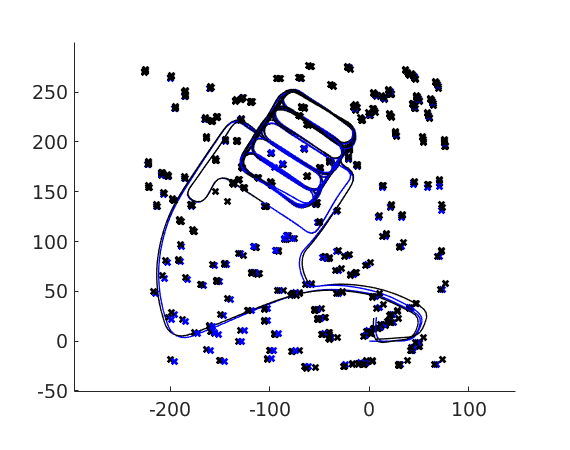}
        \caption{Parking Garage (3D)}
    \end{subfigure}%
    \newline
    \begin{subfigure}{0.5\linewidth}
        \centering
        \includegraphics[width=\linewidth]{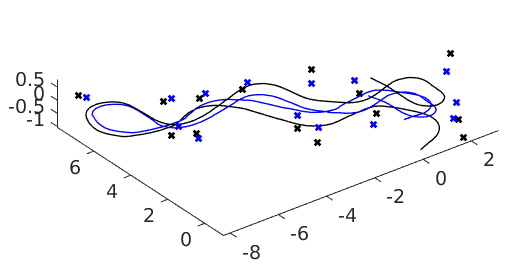}
        \caption{Indoor (3D)}
    \end{subfigure}
    \caption{Benchmarking datasets. The black trajectories and markers denote the initial guess for the trajectories (odometry) and the landmarks. The blue trajectories and markers are the optimized results given 
    ground-truth data associations.}\label{fig:mapviz}\vspace{-3mm}
\end{figure}

%% file: sections/fig-sensitivityBeta.tex
\begin{figure}[t]
    \centering
    \begin{subfigure}{\linewidth}
        \centering
        \includegraphics[width=0.49\linewidth]{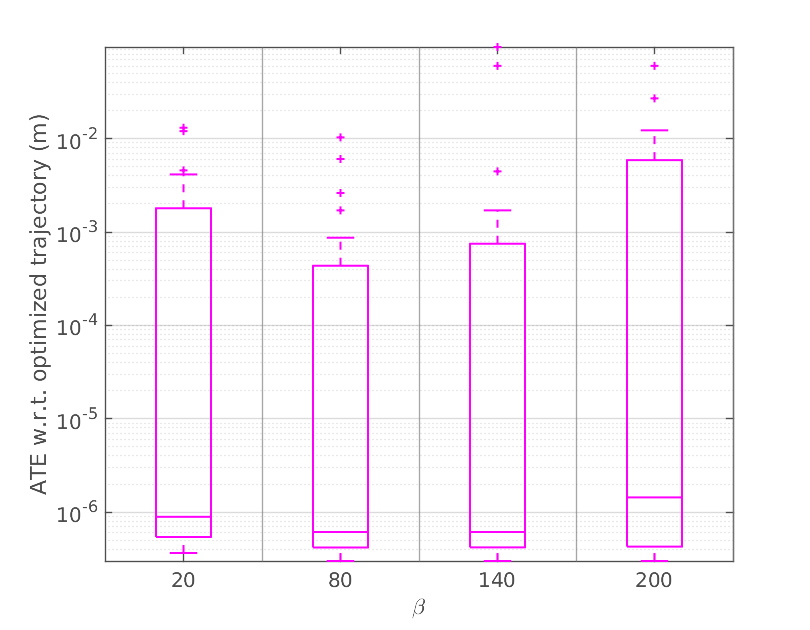}  
        \includegraphics[width=0.49\linewidth]{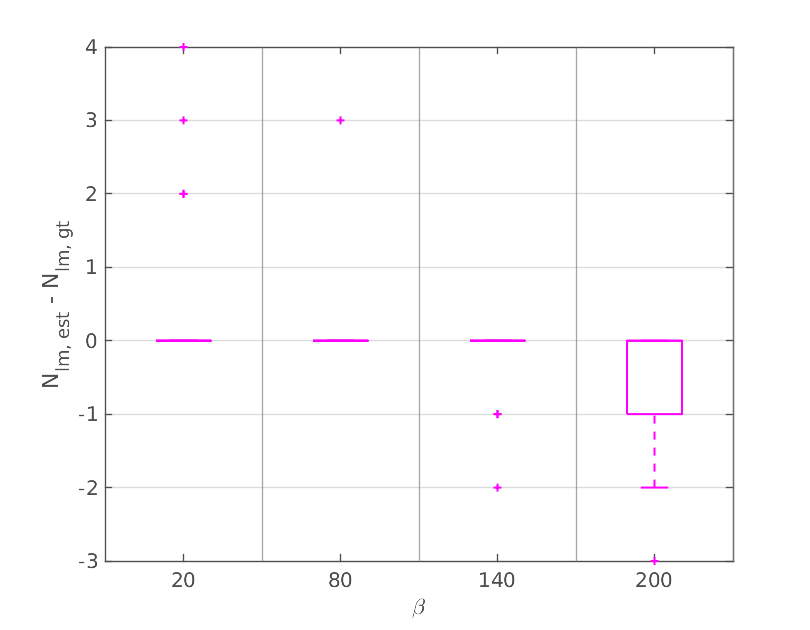}
        \caption{Default Noise Level}
    \end{subfigure}%
    \\
    \begin{subfigure}{\linewidth}
        \centering
        \includegraphics[width=0.49\linewidth]{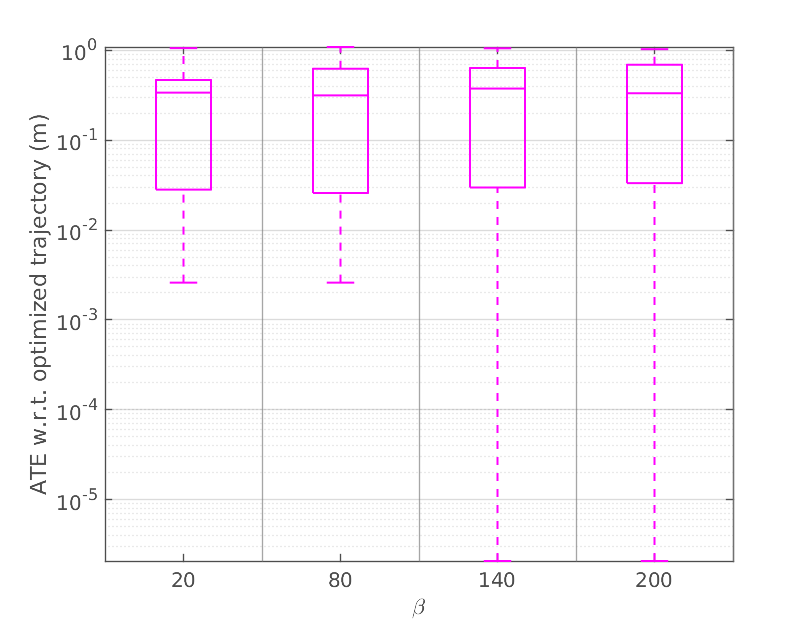}  
        \includegraphics[width=0.49\linewidth]{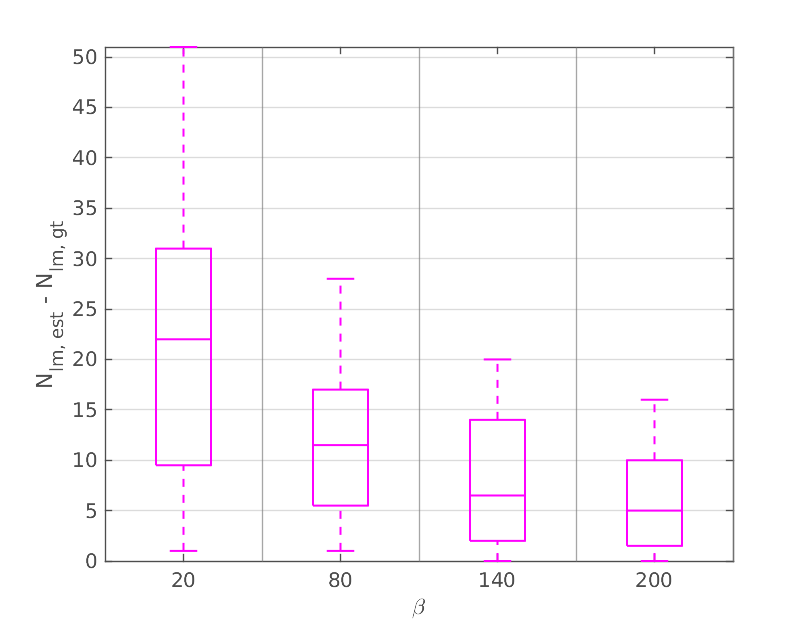}
        \caption{High Odometry Noise\label{fig:senshighodom}}
    \end{subfigure}%
    \\
    \begin{subfigure}{\linewidth}
        \centering
        \includegraphics[width=0.49\linewidth]{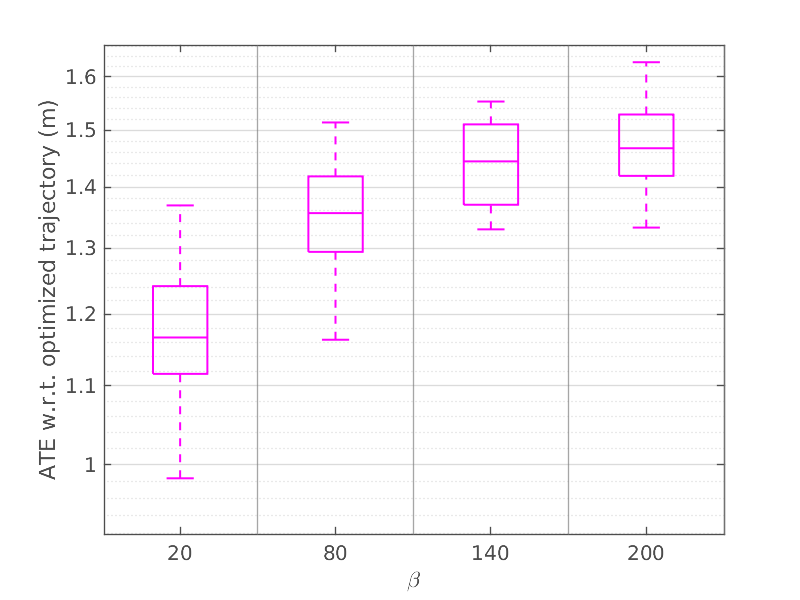}  
        \includegraphics[width=0.49\linewidth]{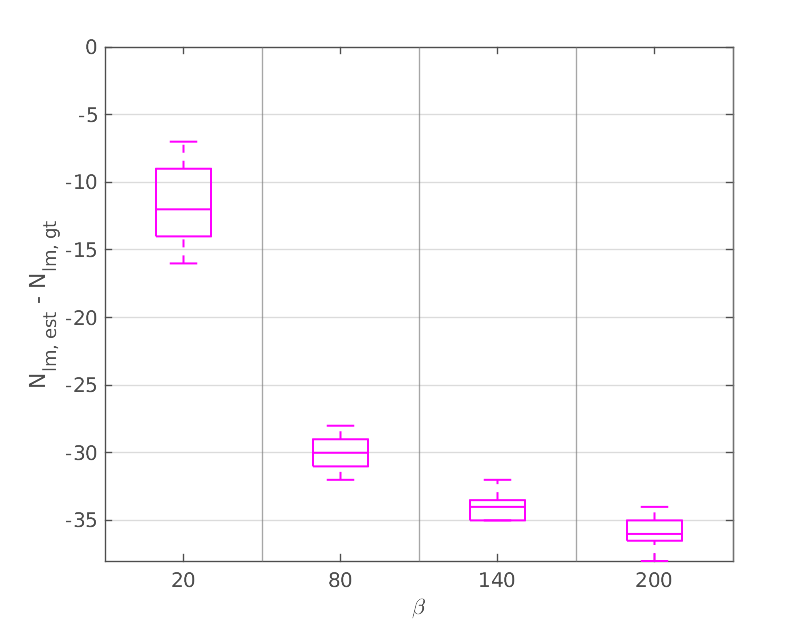}
        \caption{High Landmark Measurement Noise\label{fig:senshighldmk}}
    \end{subfigure}%
    \caption{Absolute trajectory error (ATE) and number of landmarks estimation for different values of $\beta$ in \nameKmeans on the Grid (3D) dataset. The default noise level is given in Table~\ref{tab:simparams}. The high noise cases, Fig.~\ref{fig:senshighodom} and~\ref{fig:senshighldmk}, use the maximum noise levels in Fig.~\ref{fig:odomNoise3D} and~\ref{fig:lmkNoiseAndNumber3D}, respectively.\label{fig:sensitivity}}
    \vspace{-10pt}
\end{figure}

%% file: sections/fig-ablation.tex

\begin{figure}
    \centering
    \begin{subfigure}{0.5\linewidth}
        \centering
        \includegraphics[width=1.1\linewidth]{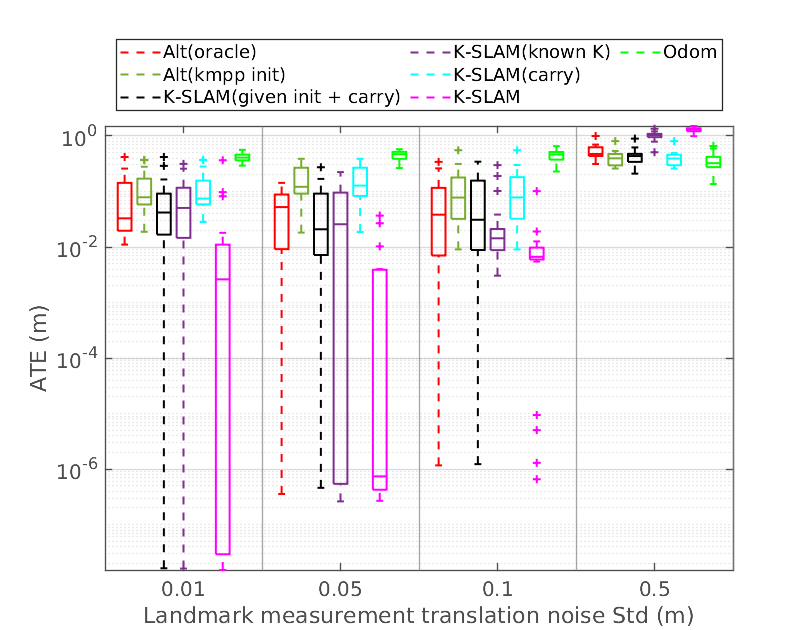}
        \caption{ATE\label{fig:ablationa}}
    \end{subfigure}%
    \begin{subfigure}{0.5\linewidth}
        \centering
        \includegraphics[width=1.1\linewidth]{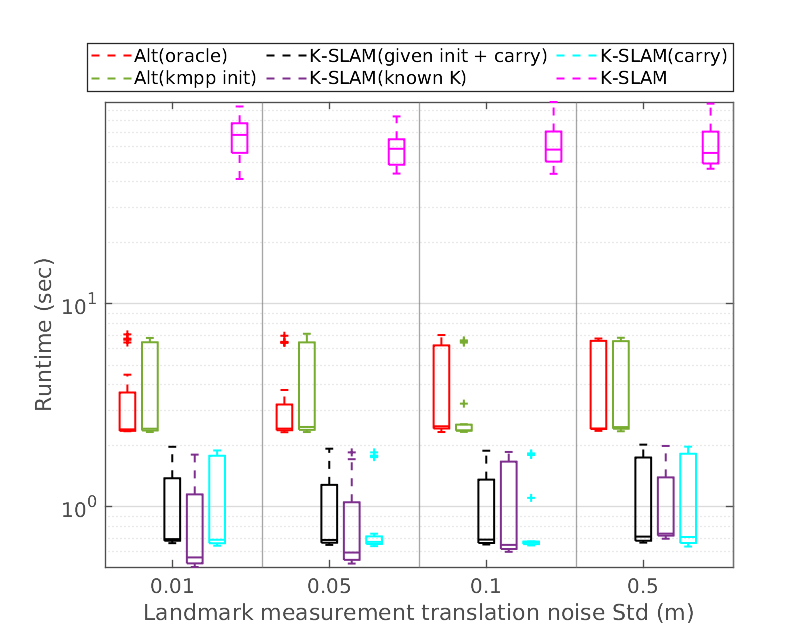}
        \caption{Runtime\label{fig:ablationb}}
    \end{subfigure}%
    \caption{Ablation study on the Grid (3D) dataset. (i) Alt (kmpp init): the oracle alternation but with the initial guess for the landmark positions computed by running k-means with k-means++~\cite{Arthur07-kmeanspp} on the projected landmark measurements (in the same way as lines~\ref{line:proj}-\ref{line:kmeans} of Algorithm~\ref{algo:inner}). (ii) K-SLAM (given init + carry): K-SLAM but given the initial guess for landmarks (same as the initial guess given to \nameAlt. $K$ is also known.) and with $\vy$ initialized by the previous iteration result (starting from iteration 2) instead of k-means++ in line~\ref{line:kmeans++} of Algorithm~\ref{algo:inner}. In other words, in the first iteration, line~\ref{line:kmeans++} in Algorithm~\ref{algo:inner} is replaced by the given initial guess and in later iterations, it is replaced by the estimate of $\vy$ from the last iteration. (iii) K-SLAM (known $K$): K-SLAM given the ground-truth $K$ (\ie simply Algorithm~\ref{algo:inner}). (iv) K-SLAM (carry): K-SLAM given the ground-truth $K$ and with $\vy$ initialized by the previous iteration result (starting from iteration 2) instead of k-means++ in line~\ref{line:kmeans++} of Algorithm~\ref{algo:inner}. \label{fig:ablation} \vspace{-4mm}}
\end{figure}

%% file: root.bbl
\begin{thebibliography}{10}

\bibitem{Agarwal13icra}
P.~Agarwal, G.~D. Tipaldi, L.~Spinello, C.~Stachniss, and W.~Burgard.
\newblock Robust map optimization using dynamic covariance scaling.
\newblock In {\em IEEE Intl. Conf. on Robotics and Automation (ICRA)}, 2013.

\bibitem{Arthur07-kmeanspp}
D.~Arthur and S.~Vassilvitskii.
\newblock {K-Means++}: The advantages of careful seeding.
\newblock pages 1027--1035, 2007.

\bibitem{bailey02thesis}
Tim Bailey.
\newblock {\em Mobile Robot Localisation and Mapping in Extensive Outdoor
  Environments}.
\newblock PhD thesis, University of Sydney, 2002.

\bibitem{bar1990tracking}
Yaakov Bar-Shalom, Thomas~E Fortmann, and Peter~G Cable.
\newblock Tracking and data association, 1990.

\bibitem{Bowman17icra}
S.L. Bowman, N.~Atanasov, K.~Daniilidis, and G.J. Pappas.
\newblock Probabilistic data association for semantic {SLAM}.
\newblock In {\em IEEE Intl. Conf. on Robotics and Automation (ICRA)}, pages
  1722--1729, 2017.

\bibitem{Cadena16tro-SLAMsurvey}
Cesar Cadena, Luca Carlone, Henry Carrillo, Yasir Latif, Davide Scaramuzza,
  Jos{\'e} Neira, Ian Reid, and John~J Leonard.
\newblock Past, present, and future of simultaneous localization and mapping:
  Toward the robust-perception age.
\newblock {\em IEEE Transactions on robotics}, 32(6):1309--1332, 2016.

\bibitem{Dellaert00cvpr-correspondenceFreeSfm}
F.~Dellaert, S.M. Seitz, C.E. Thorpe, and S.~Thrun.
\newblock Structure from motion without correspondence.
\newblock In {\em Proc. {IEEE} Int. Conf. Computer Vision and Pattern
  Recognition}, volume~2, pages 557--564, 2000.

\bibitem{doherty2020probabilistic}
Kevin~J Doherty, David~P Baxter, Edward Schneeweiss, and John~J Leonard.
\newblock Probabilistic data association via mixture models for robust semantic
  {SLAM}.
\newblock In {\em 2020 IEEE International Conference on Robotics and Automation
  (ICRA)}, pages 1098--1104. IEEE, 2020.

\bibitem{doherty2022discrete}
Kevin~J Doherty, Ziqi Lu, Kurran Singh, and John~J Leonard.
\newblock Discrete-continuous smoothing and mapping.
\newblock {\em arXiv preprint arXiv:2204.11936}, 2022.

\bibitem{gtsam}
{F. Dellaert et al.}
\newblock {Georgia Tech Smoothing And Mapping (GTSAM)}.
\newblock \url{https://gtsam.org/}, 2019.

\bibitem{fathian2020clear}
Kaveh Fathian, Kasra Khosoussi, Yulun Tian, Parker Lusk, and Jonathan~P How.
\newblock Clear: A consistent lifting, embedding, and alignment rectification
  algorithm for multiview data association.
\newblock {\em IEEE Transactions on Robotics}, 36(6):1686--1703, 2020.

\bibitem{kuhn_1955}
H.~W. Kuhn.
\newblock The hungarian method for the assignment problem.
\newblock {\em Naval Research Logistics Quarterly}, 2(1-2):83–97, 1955.

\bibitem{Kuemmerle11icra}
R.~K{\"u}mmerle, G.~Grisetti, H.~Strasdat, K.~Konolige, and W.~Burgard.
\newblock g2o: A general framework for graph optimization.
\newblock In {\em Proc.~of the IEEE Int.~Conf.~on Robotics and Automation
  (ICRA)}, May 2011.

\bibitem{kummerle2011g}
Rainer K{\"u}mmerle, Giorgio Grisetti, Hauke Strasdat, Kurt Konolige, and
  Wolfram Burgard.
\newblock g 2 o: A general framework for graph optimization.
\newblock In {\em 2011 IEEE International Conference on Robotics and
  Automation}, pages 3607--3613. IEEE, 2011.

\bibitem{g2odata}
Rainer K{\"u}mmerle, Giorgio Grisetti, Hauke Strasdat, Kurt Konolige, and
  Wolfram Burgard.
\newblock G2o datasets.
\newblock \url{https://github.com/OpenSLAM-org/openslam_g2o/tree/master/data},
  2011.

\bibitem{levenberg_1944}
Kenneth Levenberg.
\newblock A method for the solution of certain non-linear problems in least
  squares.
\newblock {\em Quarterly of Applied Mathematics}, 2(2):164–168, 1944.

\bibitem{li2021odam}
Kejie Li, Daniel DeTone, Yu~Fan~Steven Chen, Minh Vo, Ian Reid, Hamid
  Rezatofighi, Chris Sweeney, Julian Straub, and Richard Newcombe.
\newblock {ODAM}: Object detection, association, and mapping using posed {RGB}
  video.
\newblock In {\em Proceedings of the IEEE/CVF International Conference on
  Computer Vision}, pages 5998--6008, 2021.

\bibitem{Lusk21icra-clipper}
Parker~C Lusk, Kaveh Fathian, and Jonathan~P How.
\newblock {CLIPPER: A Graph-Theoretic Framework for Robust Data Association}.
\newblock In {\em IEEE Intl. Conf. on Robotics and Automation (ICRA)}, 2021.

\bibitem{marquardt_1963}
Donald~W. Marquardt.
\newblock An algorithm for least-squares estimation of nonlinear parameters.
\newblock {\em Journal of the Society for Industrial and Applied Mathematics},
  11(2):431--441, 1963.

\bibitem{mathworksdata}
Math{W}orks.
\newblock Math{W}orks indoor dataset.
\newblock
  \url{https://www.mathworks.com/help/nav/ug/landmark-slam-using-apriltag-markers.html}.

\bibitem{michael2022probabilistic}
Elad Michael, Tyler Summers, Tony~A Wood, Chris Manzie, and Iman Shames.
\newblock Probabilistic data association for semantic {SLAM} at scale.
\newblock {\em arXiv preprint arXiv:2202.12802}, 2022.

\bibitem{mullane2011random}
John Mullane, Ba-Ngu Vo, Martin~D Adams, and Ba-Tuong Vo.
\newblock A random-finite-set approach to {B}ayesian {SLAM}.
\newblock {\em IEEE transactions on robotics}, 27(2):268--282, 2011.

\bibitem{munkres_1957}
James Munkres.
\newblock Algorithms for the assignment and transportation problems.
\newblock {\em Journal of the Society for Industrial and Applied Mathematics},
  5(1):32–38, 1957.

\bibitem{mur2017orb}
Raul Mur-Artal and Juan~D Tard{\'o}s.
\newblock {ORB-SLAM2}: An open-source {SLAM} system for monocular, stereo, and
  {RGB-D} cameras.
\newblock {\em IEEE transactions on robotics}, 33(5):1255--1262, 2017.

\bibitem{neira2001data}
Jos{\'e} Neira and Juan~D Tard{\'o}s.
\newblock Data association in stochastic mapping using the joint compatibility
  test.
\newblock {\em IEEE Transactions on robotics and automation}, 17(6):890--897,
  2001.

\bibitem{Olson12rss}
E.~Olson and P.~Agarwal.
\newblock Inference on networks of mixtures for robust robot mapping.
\newblock In {\em Robotics: Science and Systems (RSS)}, July 2012.

\bibitem{april}
Edwin Olson.
\newblock Apriltag: A robust and flexible visual fiducial system.
\newblock In {\em 2011 IEEE International Conference on Robotics and
  Automation}, pages 3400--3407, 2011.

\bibitem{Rosen18ijrr-sesync}
David~M Rosen, Luca Carlone, Afonso~S Bandeira, and John~J Leonard.
\newblock {SE-Sync}: A certifiably correct algorithm for synchronization over
  the special {E}uclidean group.
\newblock {\em The International Journal of Robotics Research},
  38(2-3):95--125, 2019.

\bibitem{Sunderhauf12iros}
N.~S\"{u}nderhauf and P.~Protzel.
\newblock Switchable constraints for robust pose graph {SLAM}.
\newblock In {\em IEEE/RSJ Intl. Conf. on Intelligent Robots and Systems
  (IROS)}, 2012.

\bibitem{thorndike1953belongs}
Robert Thorndike.
\newblock Who belongs in the family?
\newblock {\em Psychometrika}, 18(4):267--276, 1953.

\bibitem{Yang20ral-GNC}
H.~Yang, P.~Antonante, V.~Tzoumas, and L.~Carlone.
\newblock Graduated non-convexity for robust spatial perception: From
  non-minimal solvers to global outlier rejection.
\newblock {\em {IEEE} Robotics and Automation Letters}, 5(2):1127--1134, 2020.
\newblock arXiv preprint arXiv:1909.08605 (with supplemental material),
  \linkToPdf{https://arxiv.org/pdf/1909.08605.pdf}\award{, ICRA Best paper
  award in Robot Vision}.

\end{thebibliography}
